\documentclass[lettersize,journal]{IEEEtran}
\usepackage{amsmath,amsfonts}
\usepackage{authblk}
\usepackage{algorithm}
\usepackage{array}
\usepackage{textcomp}
\usepackage{stfloats}
\usepackage{url}
\usepackage{verbatim}
\usepackage{graphicx}
\usepackage{cite}
\usepackage{amssymb}
\usepackage{amsmath}
\usepackage{bm}
\usepackage{subcaption}
\usepackage{graphicx}
\usepackage{algpseudocode}
\usepackage{multirow}
\usepackage{makecell}
\usepackage{svg}
\usepackage{epsfig}
\usepackage{caption}
\usepackage{subcaption}

\begin{document}

\title{Pneumatic bellows actuated parallel platform
control with adjustable stiffness using a hybrid
feed-forward and variable gain I-controller}

\author[1]{Martin Varga}
\author[1]{Ivan Virgala}
\author[1]{Michal Kelemen}
\author[1]{Ľubica Miková}
\author[2]{Zdenko Bobovský}
\author[1]{Peter Ján Sinčák}
\author[1]{Tomáš Merva}
\affil[1]{Faculty of Mechanical Engineering, Technical University of Košice, Slovakia}
\affil[2]{Faculty of Mechanical Engineering, Technical University of Ostrava, Czech Republic}

\markboth{}
{Shell \MakeLowercase{\textit{et al.}}: A Sample Article Using IEEEtran.cls for IEEE Journals}

\IEEEpubid{}

\maketitle

\begin{abstract}
Redundant cascade manipulators actuated by pneumatic bellows actuators are passively compliant, rugged and dexterous which are qualities making them exceptionally well suited for applications in agriculture. Unfortunately bellows actuators are notoriously difficult to precisely position.
This paper presents a novel control algorithm for the control of a parallel platform actuated by pneumatic bellows actuators, which is serving as one module of a cascade manipulator. The algorithm combines a feed-forward controller and a variable gain I-controller. The feed-forward controller was designed using experimental data and two regression steps to create a mathematical representation of the data. The gain of the I-controller depends linearly on the total reference error, which allows the I-controller to work in concert with the feed-forward part of the controller. The presented algorithm was experimentally verified and its performance was compared with two controllers, an ANFIS controller and a constant gain PID controller, to satisfactory results. The controller was also tested under dynamic loading conditions showing promising results. 
\end{abstract}


\section{Introduction}
\IEEEPARstart{I}{ndustrial} robots are an indispensable part of the manufacturing process in many parts of the industry, where their traits, i.e., precision, speed, and the ability to work basically nonstop, help increase productivity and decrease cost. It is therefore understandable, that there is a strong incentive to use industrial robots in other fields, like agriculture and medicine, to name a few. The most common industrial robots are serial link 6R robots, 2R1T SCARA robots or parallel Delta robots driven by, most commonly, electric actuators, or in some cases hydraulic actuators \cite{2022}. These industrial robots were developed for many decades, their design is standardized, and their mathematical description and control design is fairly well researched. Unfortunately, these robots lack some key features needed in the aforementioned new fields of application. For example compliance, agility and complex modes of motion \cite{2018}. These requirements fulfil new emerging classes of robots i.e., redundant cascade and continuum robots \cite{VIRGALA2014489}. The redundant robots are all those that have more degrees of freedom than is necessary to perform a certain task. Development of these robots took pace from the year 2000. Redundant robots in general, but especially cascade and continuum robots, have unique characteristics. For example, according to \cite{174447} and \cite{388263}, compliance, a good reach to weight ratio, modularity and other. These characteristics arise from their specific design. 

In general, continuum and cascade robots consist of several in series connected parallel modules that, if underactuated, form a continuum robot and if fully actuated form a cascade robot. To describe the motion of a redundant robot and design a control algorithm, it is first necessary to focus on its individual modules and their properties. The chosen actuator type influences the achievable properties of a module. Electrical linear servo motors as used by \cite{Zhao2020} within the structure of a module or outside as done by \cite{9764135} provide high force, stiffness and are usually equipped with position sensors simplifying control. Hydraulic actuators as can be seen in \cite{PI2011185} provide high forces and can be precisely positioned, jet are slow. Nonstandard actuators for such applications can also be used, for example SMA springs as seen in the work of \cite{8730312} or dielectric materials as described by \cite{Wingert2002HyperredundantRM}. Pneumatic actuators are a popular class of actuators that are used in these applications. They provide high power density, relatively low weight, can be easily manufactured to custom specification, as described in \cite{2022} and \cite{2019}, or bought off the shelve in a variety of types and sizes \cite{Tao2016}, \cite{automation20164210}, \cite{7049581} and the compressibility of air give them a natural level of compliance. This ability makes them the actuator of choice for medical applications, like rehabilitation equipment \cite{Wang2021}, in flexible endoscopes \cite{6696866}, \cite{8333287}, in agriculture \cite{mrx05}, as parts of  mobile robots \cite{DBLP:journals/corr/abs-2108-01338}, \cite{10.1007/3-540-26415-9_89},  as the actuator for a high precision positioning system \cite{Fujita_2011ijat} or a stiffness regulating element in hybrid actuation schemes for continuum tendon driven robots as presented by paper \cite{HARSONO2022105067}. 

Control of a parallel platform module actuated by three or four pneumatic actuators is a challenging task. One approach to control of these modules is to use a feed-forward controller alone or in combination with other types. \cite{9790324} presented a modelling framework to design a model from which a feed-forward controller with satisfactory performance was developed. The work \cite{9046840} developed a custom bellows type actuator and applied it in a parallel platform with three degrees of freedom. The control algorithm was a feed-forward controller based on experimental mapping between pose, external forces and input pressure. In \cite{6347932} authors used a feed-forward controller based on a mathematical model in combination with a variable P gain PI controller to combat the effect of hysteresis in the system. To positioning of a soft robot, the paper \cite{8722799} uses a model based controller using both feed-forward and feedback components with a structure similar to a PD controller. A planar platform actuated by pneumatic muscles is controlled by \cite{KHOA2013462} using three fuzzy controllers synchronized through an ANFIS (Adaptive neuro fuzzy inference system) based controller. The paper \cite{7049581} applied a simple constant gain PID controller for positioning of a parallel platform actuated by four pneumatic muscles. \cite{9088970} have demonstrated a nonlinear SMC (sliding mode controller) based on a PID type sliding surface combined with a lumped element model-based controllers to control a soft pneumatically actuated robot. In the paper \cite{2011} authors uses a fourier series-based adaptive sliding-mode controller with $H\infty$ tracking performance to solve the high non-linearity and time-varying problem for a parallel platform actuated by rod-less pneumatic cylinders. While \cite{Mo2022} dealing with a tendon driven redundant manipulator proposes a population-based model-free control method that could be applied to pneumatically actuated manipulators as well. 

Based on previous works we have set up to develop the mechanical design and control system for a rugged redundant cascade manipulator driven by pneumatic bellows intended for both research and agricultural use. The most difficult part of the development is the controller design for one separate module, as noted in previously mentioned articles, this task is notoriously difficult due to the inherent nonlinear hysteresis behaviour of the chosen actuator type and MIMO system as a whole. A similar route to previous research in pneumatic parallel platform module control was taken by relying on experimental data, but this concept was expanded upon by applying two regression steps to the data to get a mathematical module model, which is taking the place of a feed-forward controller. This feed-forward controller was later supplemented by a variable gain I-controller that facilitates disturbance rejection. The controller design allows for a on demand change in stiffness of the system during operation and lends itself well to be a part of complete control system for the whole cascade manipulator. 
\\
Based on previous papers survey, the novelty of the paper can be defined as follows:
\begin{itemize}
    \item Development of a novel hybrid FFvI controller
    \item Establishment of controller design methodology for cascade redundant robots
    \item Experimental positioning analyses under dynamic disturbance effects 
\end{itemize}

\begin{figure}[H]
    \centering
    \includegraphics[angle=90]{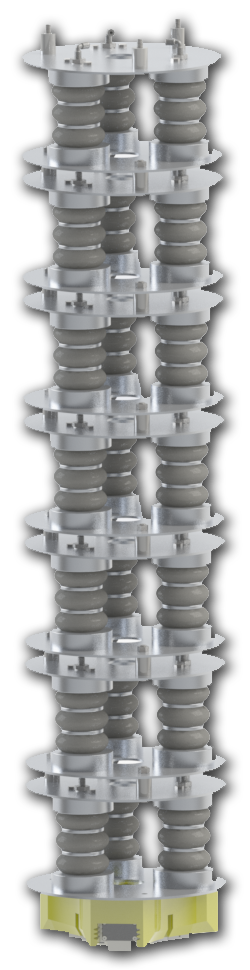}
    \caption{CAD model of PneuTrunk}
    \label{fig:pneuTrunk}
\end{figure}

This paper focuses on the development of a novel controller for a 2 DOF pneumatic parallel platform that represents one module of the pneumatic manipulator PneuTrunk (see Fig. \ref{fig:pneuTrunk}), developed by our Cognitics Lab. The first part of the paper presents the design and kinematic model of one module of PneuTrunk. In the second part a feed-forward controller based on an experimentally identified system with stiffness regulation capabilities combined with a variable gain I controller for disturbance rejection is presented. In the third part, the proposed controller is compared with a simple PID controller and ANFIS controller.

\section{Design of PneuTrunk module}
\begin{figure}[H]
    \centering
    \includegraphics[trim={2cm 4cm 6cm 4.5cm},clip, width=80mm]{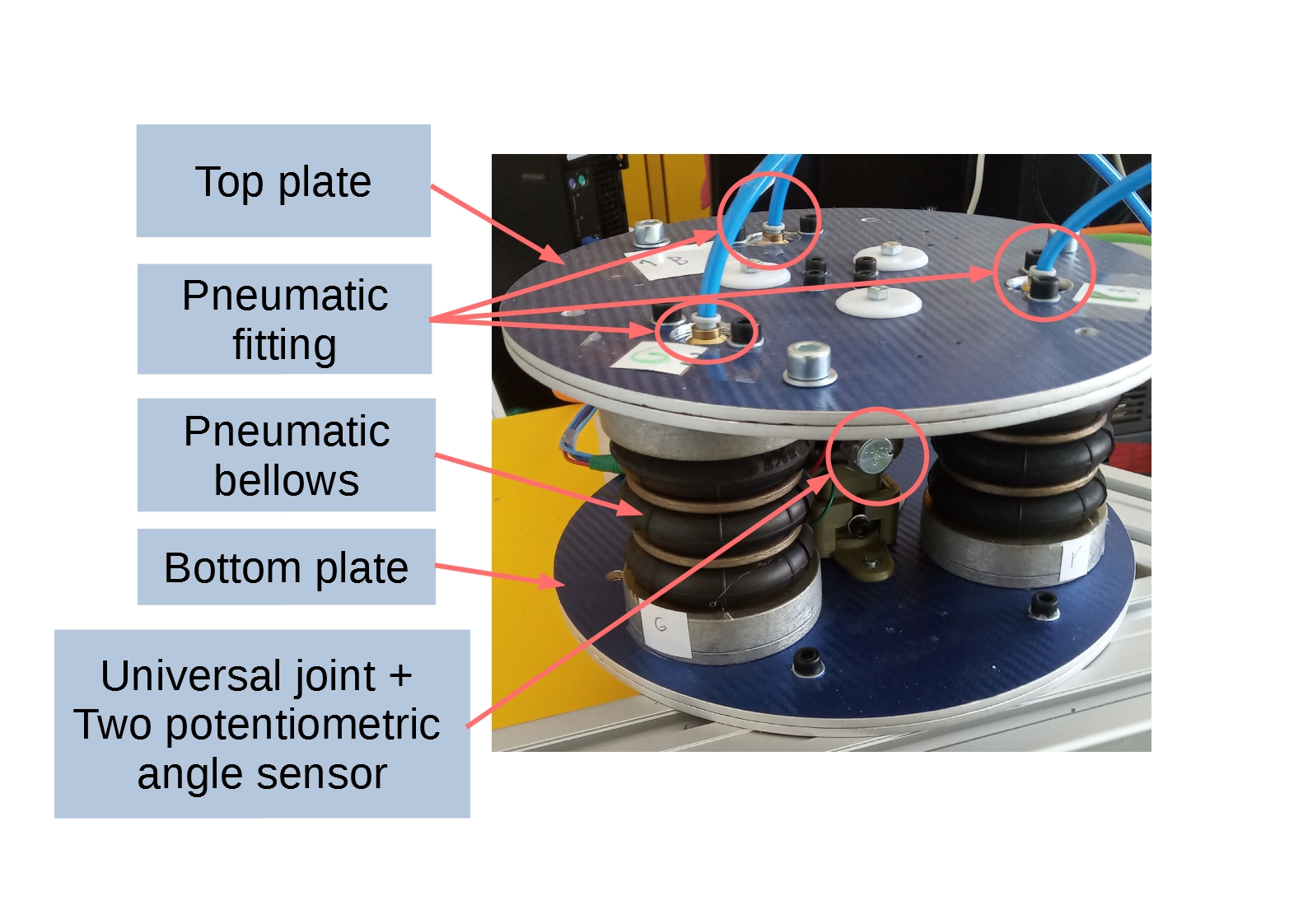}
    \caption{One module of the manipulator PneuTrunk}
    \label{fig:oneModule}
\end{figure}

As can be seen in Fig. \ref{fig:pneuTrunk}, the redundant manipulator PneuTrunk is a cascade type manipulator constructed out of parallel platform modules ordered in series. The number of modules depends on the required degrees of freedom. One module, shown in Fig. \ref{fig:oneModule}, consists of two duraluminium plates connected by an universal joint and three evenly spaced pneumatic bellows. The tilt angles between the top and the bottom plate are measured by two potentiometric rotation sensors placed in such a way that the axis of the universal joint are colinear with both axes of the sensors. The pneumatic actuators are of the shelve Dunlop 2 3/4 x 3 bellows. The pressure in the pneumatic bellows is controlled by three separate electropneumatic converters SMC ITV1050-31F20. All tubing is of inner diameter 6 mm to eliminate the effects of tubing diameter on the dynamic behavior of the bellows. The module is controlled by a B\&R PLC type 4PPC70-0702-20B.

The maximum operating pressure for one bellow is 7 bar, but to prevent damage to the system and especially the universal joint, the allowable pressure range is set to 0-5 bar. The tilt around the $x$-axis and $y$-axis at this pressure range is written in Tab. \ref{tab:tab1}. The $x$-axis is oriented towards the center of one bellows. 

\begin{table}[H]
    \caption{Tilt extremes}
    \centering
    \begin{tabular}{|| c c c ||}
    \hline
        & tilt & tilt\\
        \hline\hline
        axis & min & max \\ 
        x & -24.4 & 16.6 \\  
        y & -21.2 & 20 \\
        \hline
    \end{tabular}
    \label{tab:tab1}
\end{table}

It is expected that the module will be driven only by positive pressure. This has important implications when designing the control algorithm for such a device. While extension of one bellows is facilitated by simply supplying pressure, compression is achieved by applying external forces coming predominantly from the extension of one, or both remaining bellows. This fact is also the reason why the minimum number of pneumatic bellows is three. Coincidentally, because the module only has two degrees of freedom, this design is inherently overactuated. This causes one posture of the module to be reachable by an infinite number of bellows input pressure combinations and, in theory, giving the system the ability to change its stiffness without changing the posture. This was taken into account when designing the control algorithm for one module.

\begin{figure}[H]
    \centering
    \includegraphics[trim={2cm 15cm 8cm 55cm},clip, width=70mm]{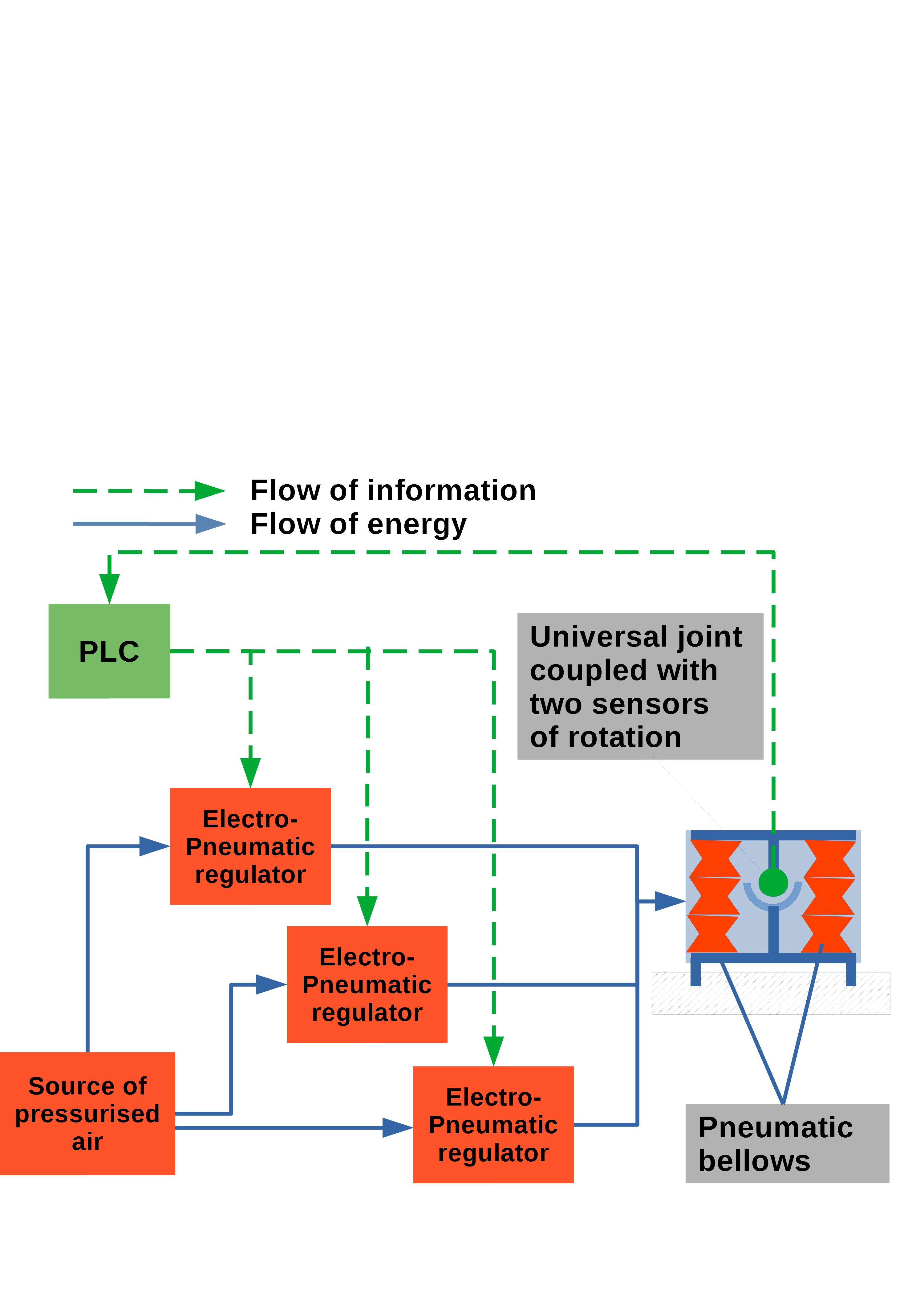}
    \caption{Flow of information and energy}
    \label{fig:flow}
\end{figure}

The flow of information and energy is visualized in Fig. \ref{fig:flow}. The whole system is simple and contains only necessary components. It could be argued that adding a center closed 2/2 valve between each bellows and its corresponding electropneumatic converter could give the system the ability to pneumatically lock the bellows extension, improving the systems positioning performance. Unfortunately, this would also complicate the system and its regulation, introducing other challenges and distracting from the aim of this paper. 

\section{Mathematical modeling}
\subsection{Mathematical model of one module}
\begin{figure}[H]
    \centering
    \includegraphics[trim={1cm 40cm 2cm 10cm},clip, width=70mm]{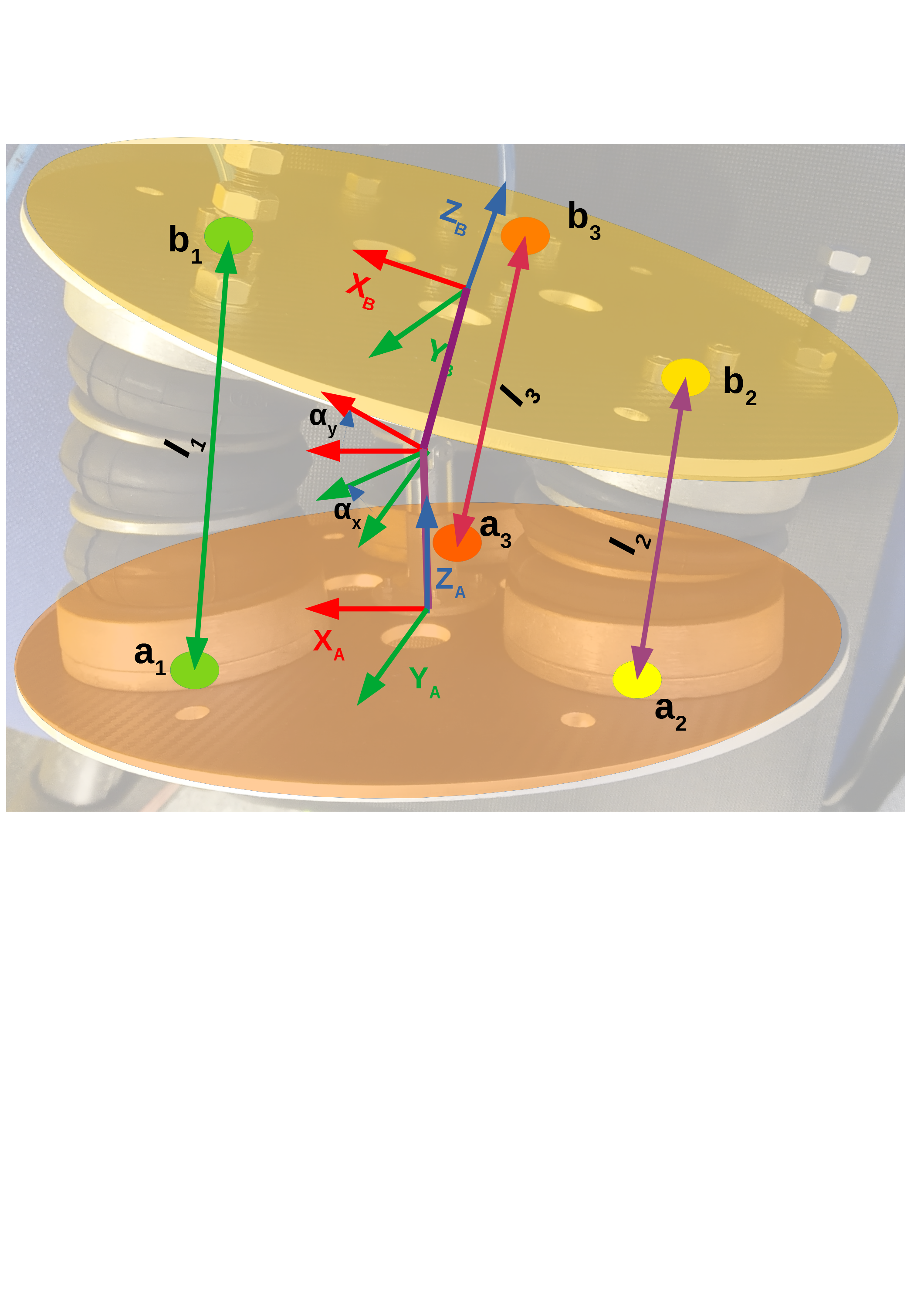}
    \caption{Schematic representation of the pneumatic module}
    \label{fig:ik}
\end{figure}

An important step before attempting to design a controller for a module is, first to create an inverse kinematic model of the module. In other words, finding a way to map the desired output parameters, here the tilting angles, to the input parameters, in this case the extension/contraction of the bellows, see Fig. \ref{fig:ik}. Inspiration is taken from the work of \cite{6878268}, where bellows type actuators are represented by two elements connected by a translational joints and connected to the bottom and top plate by universal joints. This approach greatly simplifies kinematic modelling. For the purpose of modelling the dynamics of an actuator, the model needs to be augmented by adding torque on both universal joints that represents resistance to bending of the actuator.

\begin{figure}[H]
    \centering
    \includegraphics[trim={0.13cm 0.2cm 0.2cm 0cm},clip, width=88mm]{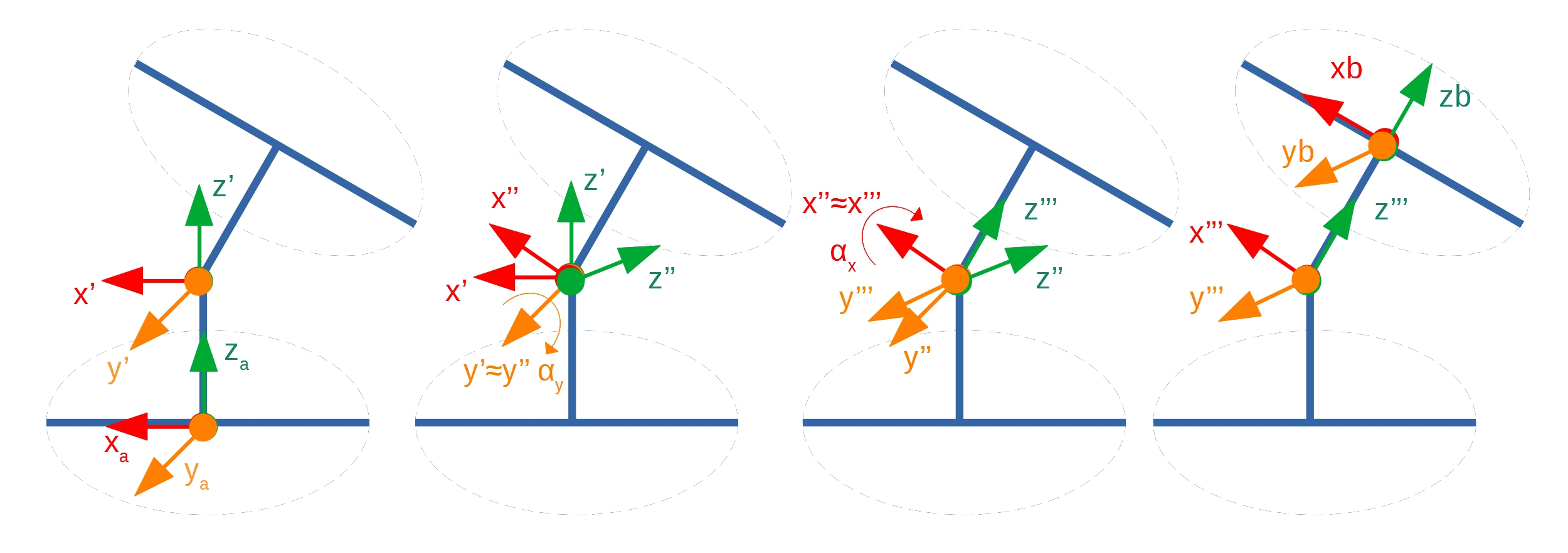}
    \caption{Transformation steps between frame $x_ay_az_a$ and $x_by_bz_b$}
    \label{fig:transf}
\end{figure}

For our design, there exists a closed form solution to inverse kinematics in the form of 
\begin{equation}
    l_{i} = \mid \textbf{a}_{i} - \textbf{H}_t\textbf{b}_{i(\alpha_{x}, \alpha_{y})}\mid
\end{equation}

\begin{equation}
    \textbf{H}_t=\textbf{T}_{z34}\times\textbf{R}_{x23}\times\textbf{R}_{y12}\times\textbf{T}_{z01}
\end{equation}
where $\emph{i} \in \{1, 2, 3\}$ denotes the bellows, $\textbf{a}_{i} \in \mathbb{R}^3$ is coordinates of the center of the bellows on the bottom plate, $\textbf{b}_{i} \in \mathbb{R}^3$ is coordinates of the center of the bellows on the top plate, $\emph{l}_{i}$ is distance between point $\textbf{a}_{i}$ and $\textbf{b}_{i}$. Matrix $\textbf{H}_{t} \in \mathbb{R}^{4 \times 4} $  represents the transformation matrix between fixed coordinate frame $x_ay_az_a$ and top plate coordinate frame $x_by_bz_b$. $\textbf{T}_{z01}, \textbf{T}_{z34}, \textbf{R}_{x23}, \textbf{R}_{y12}\in \mathbb{R}^{4 \times 4} $ where $\textbf{T}_{z01}$ is the translation matrix between the base frame and a parallel but offset frame $x'y'z'$, $\textbf{R}_{y12}$ is the rotation matrix rotating frame $x'y'z'$ around its y axis by $\alpha_{y}$ into $x''y''z''$, $\textbf{R}_{x23}$ is the rotation matrix rotating frame $x''y''z''$ around its x axis by $\alpha_{x}$ into $x'''y'''z'''$  and $\textbf{T}_{z34}$ is the translation matrix between the frame $x'''y'''z'''$ and a parallel but offset top plate frame $x_by_bz_b$, see Fig.~\ref{fig:transf}. Angle $\alpha_{x}$ is tilt angle of the top plate around axis \emph{x} and $\alpha_{y}$ is tilt angle of top plate around axis \emph{y}.

The role of the established kinematic model in relation to the tilt control of the module described in later chapters is a central one. The end goal of the control is to control the tilt, but this is achieved indirectly by controlling the extension and total pressure in the respective bellows. The presented inverse kinematic model converts the reference tilt and actual tilt sensed by rotation sensors to the required extension/contraction and actual deformation of the bellows.

\subsection{Model of pneumatic bellows}
\begin{figure}[H]
    \centering
    \includegraphics[trim={0cm 0cm 0cm 0cm},clip, width=90mm]{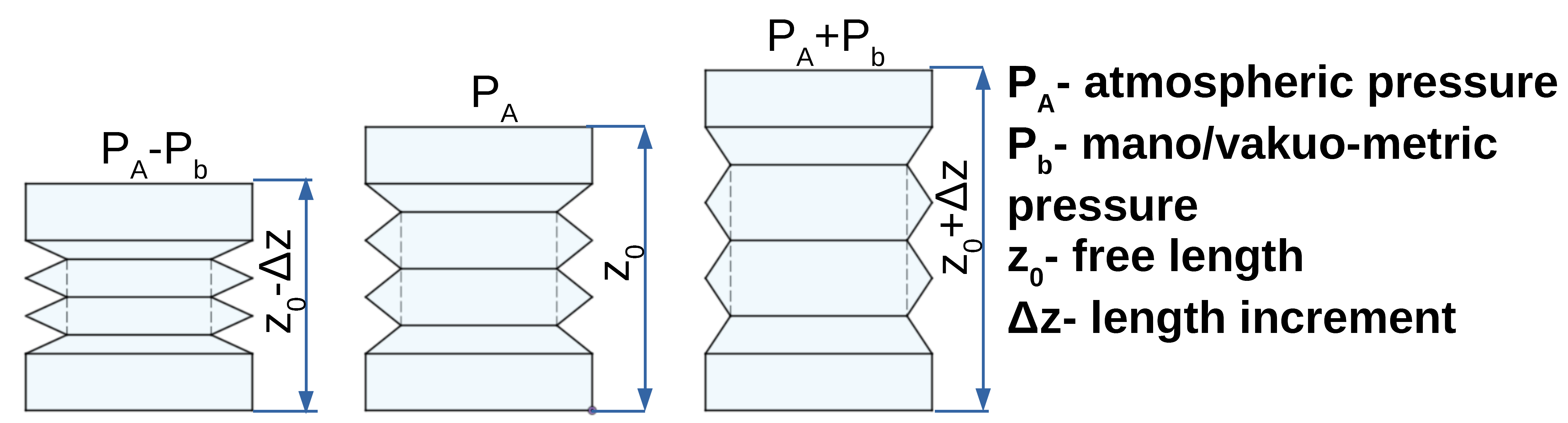}
    \caption{Bellow length in dependence on internal pressure}
    \label{fig:airSpring}
\end{figure}

A pneumatic bellows is a linear pneumatic actuator consisting of a bellows type body and mounting flanges whose free length is dependent on the difference between the ambient pressure and the pressure inside the bellows. From a physical point of view, the pneumatic bellows is a pneumatic spring with variable equilibrium length, dependent on the passive properties of the bellows and the internal pressure within the bellows Fig. \ref{fig:airSpring}.

\begin{figure}[H]
    \centering
    \includegraphics[trim={0cm 0cm 0cm 0cm},clip, width=55mm]{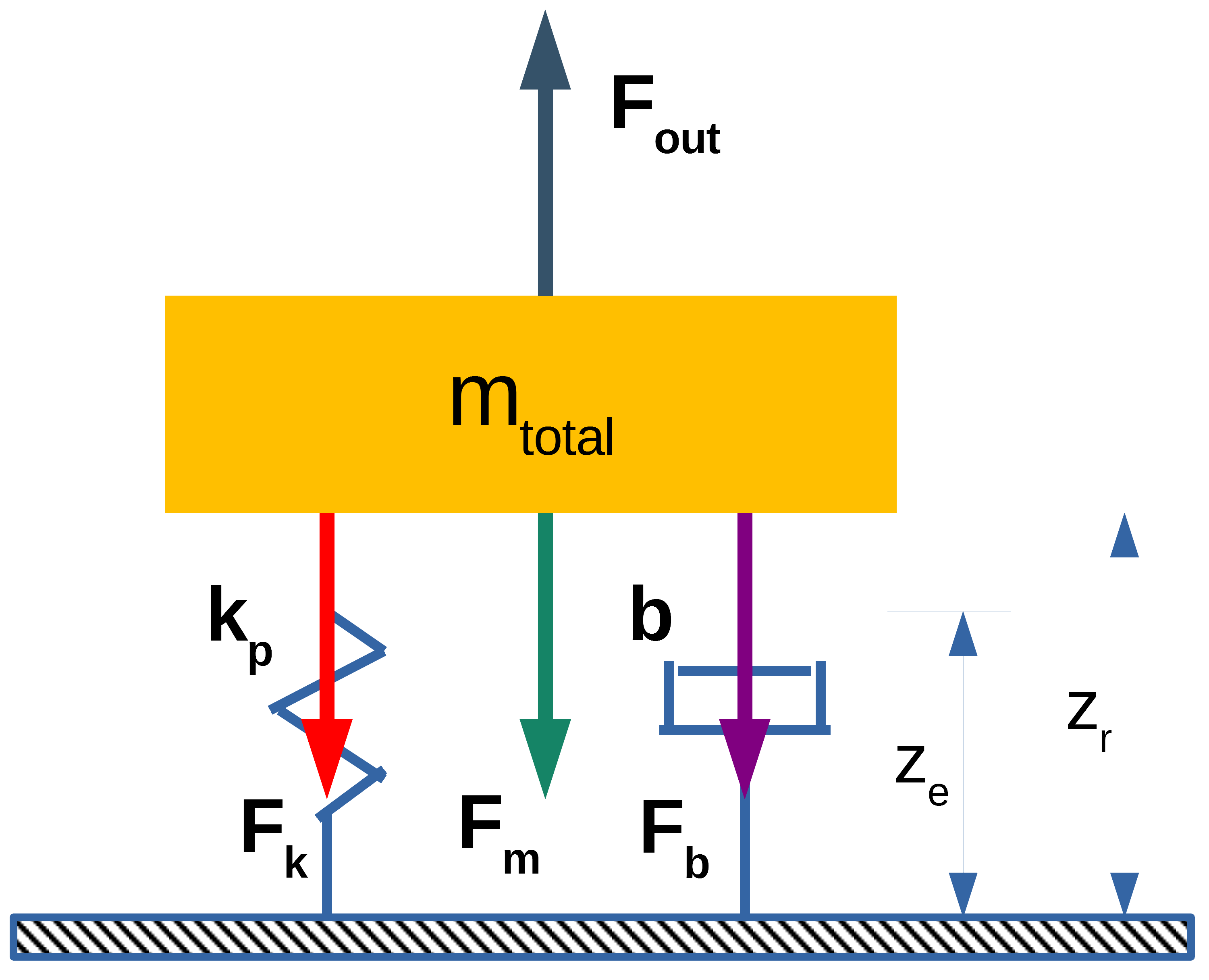}
    \caption{Mass-spring-damper model of pneumatic bellow}
    \label{fig:freeBody}
\end{figure}

\begin{figure}[H]
     \centering
     \includegraphics[trim={0cm 0cm 0cm 0cm},clip, width=55mm]{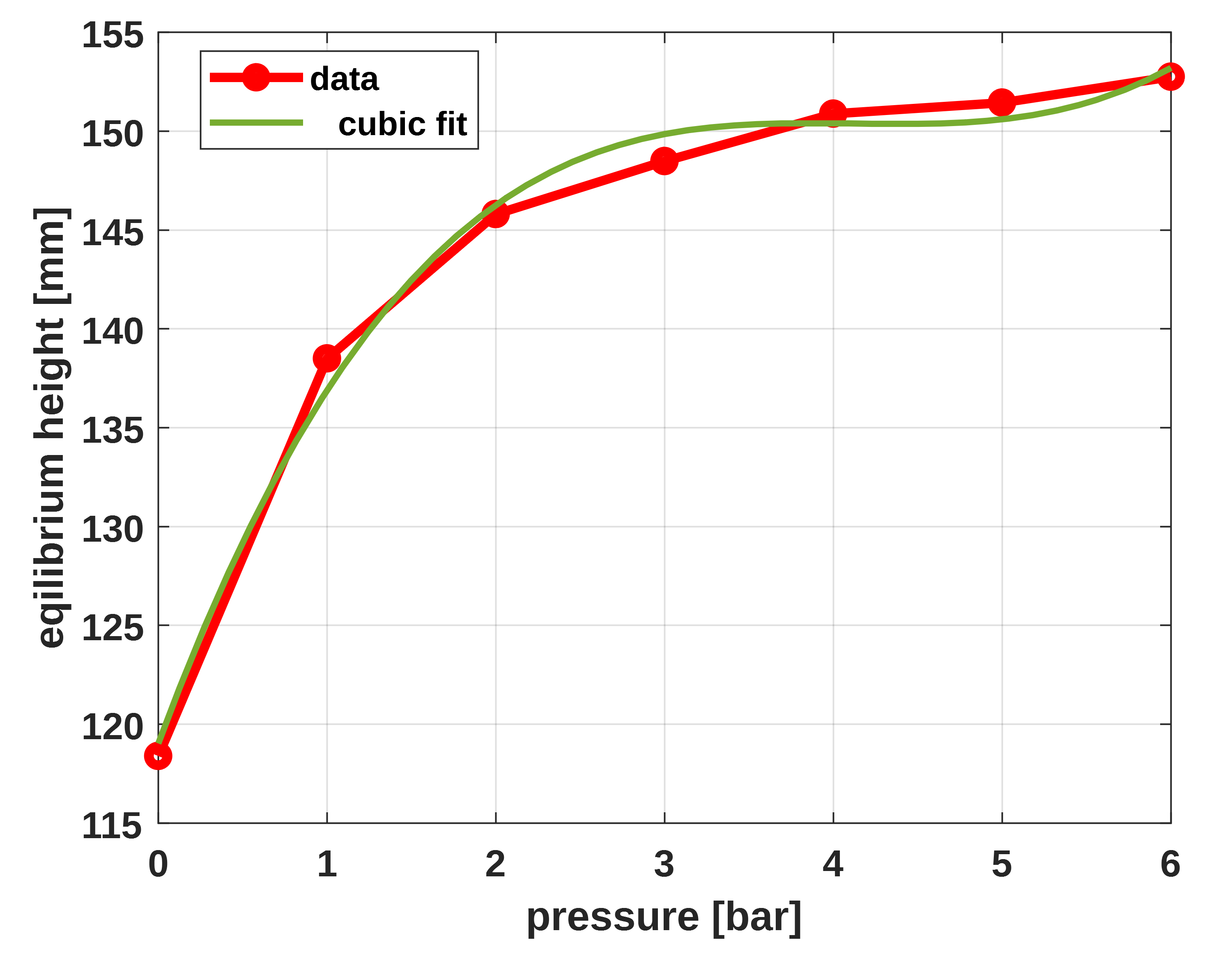}
     \caption{Relationship between internal pressure and Equilibrium height}
     \label{fig:airSpringExperimental}
 \end{figure}

The bellows can be modelled using the standard mass, spring, damper model represented by eq. \ref{eq2}, see Fig. \ref{fig:freeBody}. The dominant effect on the system has spring force and it in turn is dependent on the pneumatic spring stiffness and the equilibrium height of the bellows at the given internal pressure. The eq. \ref{eq3} shows this relationship. The bellow without being pressurized behaves like a spring whose stiffness depends on the shape of the bellow, the current material properties that are also dependent on other factors like ambient temperature. Therefore, if the bellows is deformed a spring force  appears in the direction of free height. The equilibrium length is the length of the bellows at which the deformation force from the internal pressure  is at equilibrium with the spring force. It can be seen that, the equilibrium height is a nonlinear parameter that depends on multiple other coupled parameters. Therefore, instead of a physical modelling approach, the model for the equilibrium height was derived from experimental data by measuring the equilibrium height at different internal pressures, see Fig. \ref{fig:airSpringExperimental}. The data was then interpolated by a third order polynomial function resulting in eq. \ref{eq4}.

\begin{equation}\label{eq2}
    F_M + F_b  + F_k(k_p,P_b) + F_o = 0
\end{equation}
\begin{equation}\label{eq3}
    F_k = k_p\Bigl(z_e\bigl(F_k, P_b\bigr) - z_r\Bigr)
\end{equation}
\begin{equation}\label{eq4}
    z_e = 0.45P_{b}^3 + 5.6P_{b}^2 + 23P_b +1200
\end{equation}
where $F_M$ is inertial force, $F_b$ is damping force, $F_k$ is pneumatic spring force, $F_o$ is outside disturbance force, $k_p$ is pneumatic spring stiffness, $P_b$ is internal pressure, $z_e$ is equilibrium length, $z_r$ is actual length and $F_m$ is material spring force.

 To create a simulation model of a pneumatic spring the stiffness of the spring is necessary. This parameter can be derived from the eq. \ref{eq5}, \ref{eq6} and \ref{eq7}.
 \begin{equation}\label{eq5}
     k_p = \frac{dF_k}{dz}
 \end{equation}
 where
  \begin{equation}\label{eq6}
     F_k = (P_0 - P_A)A
 \end{equation}
 Assuming that the change in bellows internal volume is polytropic we get
  \begin{equation}\label{eq7}
     P_0V^2 = constant
 \end{equation}
 Combining the above equations
  \begin{equation}\label{eq8}
     k_p = \frac{P_0nA^2}{V} + P_B\frac{dA}{dz}
 \end{equation}
 where $V$ is total volume of air within the bellows and corresponding pneumatic tube,$P_0$ is the absolute pressure inside the bellows,$A$ is an effective surface of the bellow, $n$ is polytropic constant (for this process $n = 1$).
According to eq. \ref{eq2} - \ref{eq8} a simulation model in MATLAB was developed. 

\begin{figure*}
     \centering
     \includegraphics[trim={0cm 0cm 0cm 0cm},clip, width=160mm]{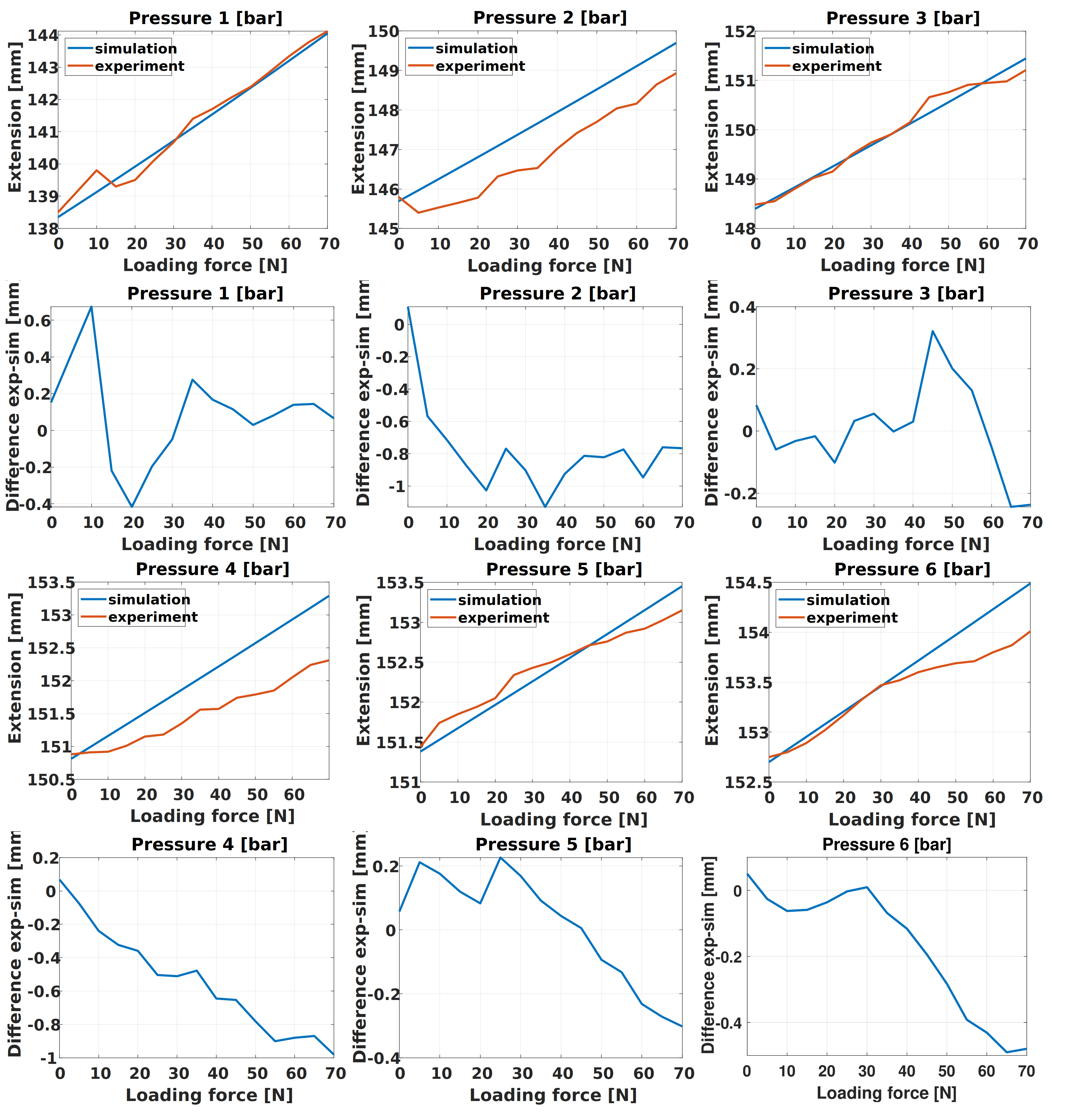}
  
\caption{Experiment vs. simulation  (1\textsuperscript{st} and 3\textsuperscript{rd} row) and the difference between simulation and experiment  (2\textsuperscript{nd} and 4\textsuperscript{th} row)}
\label{fig:sim_vlnovec}
\end{figure*}

The results of this model were compared with experimental data, where the bellow was pressurized to different pressures, a positive extension force was applied to the bellow and the total extension was measured. The results are in Fig. \ref{fig:sim_vlnovec}.

The model of pneumatic bellow gives satisfactory results. The maximum deviation for the pressures 1 bar to the pressure 6 bar does not exceed 1 mm while for pressure 0 bar the deviation is nearly 4 mm, which points to either to a measurement error or to some unknown effect that is much less pronounced in higher pressures.

\begin{figure}[H]
    \centering
    \includegraphics[trim={0cm 0cm 0cm 0cm},clip, width=55mm]{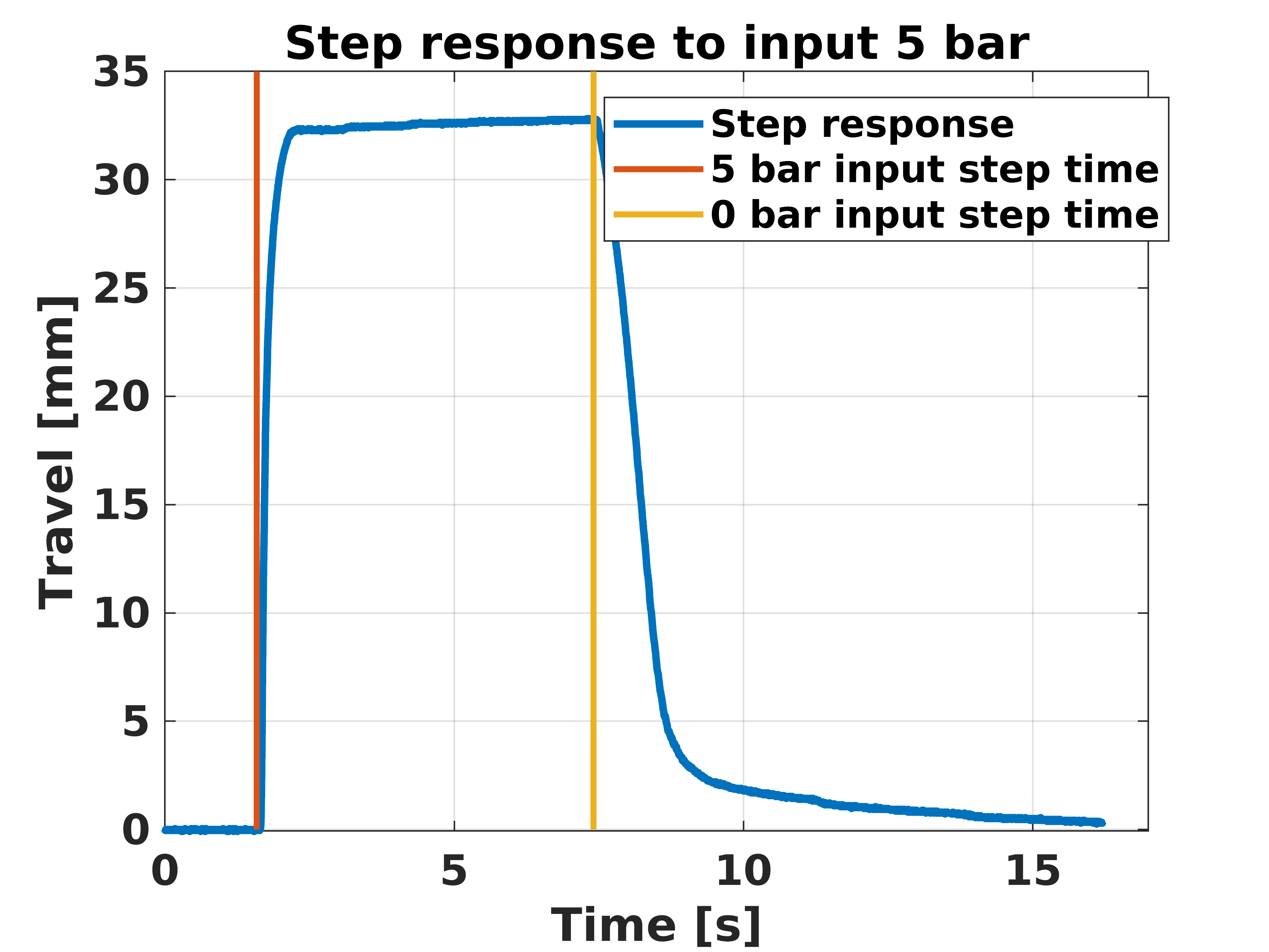}
    \caption{Step response of one bellows to step input 5 bar}
    \label{fig:stepResponse}
\end{figure}

This model represents the static behavior of an air bellow performing linear deformation. It does not capture its bending behavior or its dynamics. To be able to design a controller for one module of the manipulator PneuTrunk, it is necessary to also have a basic understanding of the dynamic behavior of one bellow. This can be seen in the step response of one bellow to an input pressure step of 5 bar, shown in Fig. \ref{fig:stepResponse}. 

There is no discernible overshoot and the rise time from 0 s to maximum value is about 0.4 s. This means that the system is overdamped and 0.4 s represents the maximum possible regulation speed. It is also important to note the behavior of the bellows when going back from 5 bar to 0 bar, where the actuator is passively returning to the original length. Here, not even after 8 s does the actuator reach the original length.

One important property of a pneumatic bellow, as noted by \cite{MISHRA2022104841}, is its hysteresis behavior, where inflating and deflating a bellow results in a different free length at zero internal pressure. In our experiments, this behavior resulted in a deviation of ± 2mm. To combat this effect, the bellow was forced by an external stop to always be extended at zero internal pressure securing a stable free length.
Creating a comprehensive bellow model falls outside the scope of this paper and will be a topic of further research. Non the less, it gives important insights into the behavior of one bellow regarding controller development.

\section{Controller design}

Controlling the posture of one module requires the combined effort of all three of its bellows actuators. The presented controller is designed to deal both with the non-linearity of the actuators and the over-actuation of the system. We define the controller consisting of two parts, a feed-forward controller and a variable gain I-controller (FFvI). 

The feed-forward control is widely used in research for these applications, for example \cite{9046840} and \cite{9790324}. It uses an inverse model of a controlled system without a feedback loop. For this application, it will provide the rough estimate input. This leverages the lack of overshoot of the actuators even at large input pressure steps, as can be seen in Fig. \ref{fig:stepResponse}, it maximizes the controller speed, and it is generally easy to design and implement. On the other hand, because of its lack of a feedback loop, as seen in \cite{9790324}, it is unable to compensate for disturbance forces and system-model deviation. These are the reasons why the feed-forward controller is supplemented by a variable gain I-controller designed to complement the feed-forward controller and dynamically react to any differences between the reference values and actual values of the controlled variables. 

\subsection{Feed-forward controller design}

\begin{figure}[h!]
    \centering
    \includegraphics[trim={0cm 0cm 0cm 0cm},clip, width=55mm]{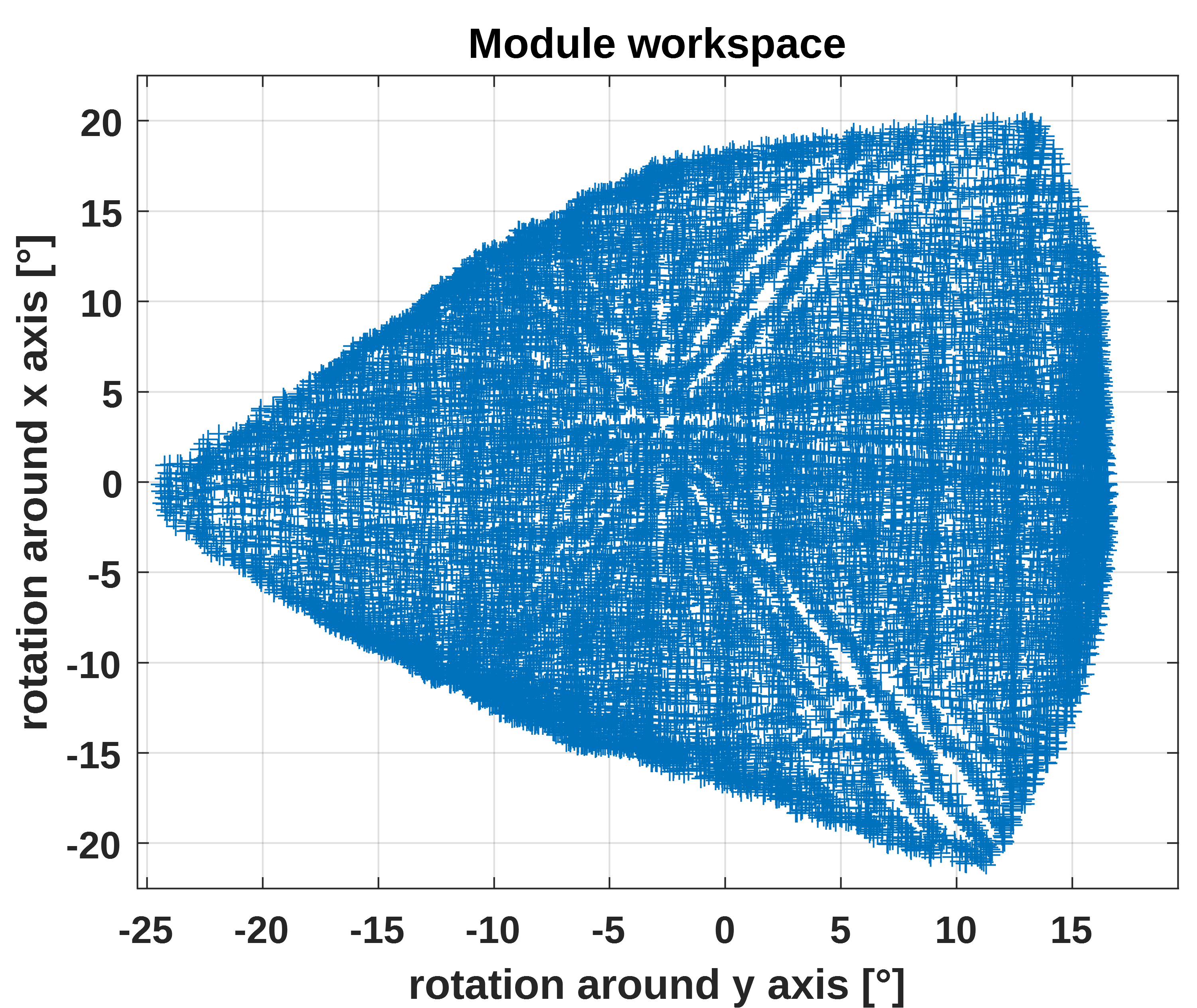}
    \caption{Module workspace}
    \label{fig:workspace}
\end{figure}

The feed-forward controller developed in this paper was designed using experimental data. Various pressure combinations were supplied to each bellows and the resulting tilt was measured. The supplied pressures ranged from 0 bar to 5 bar with a 0.2 bar increment. This results in 18275 different pressure combinations and their corresponding tilting angles. The module workspace can be seen on Fig. \ref{fig:workspace}.

\begin{figure}[h!]
    \centering
    \includegraphics[trim={0cm 0cm 0cm 0cm},clip, width=55mm]{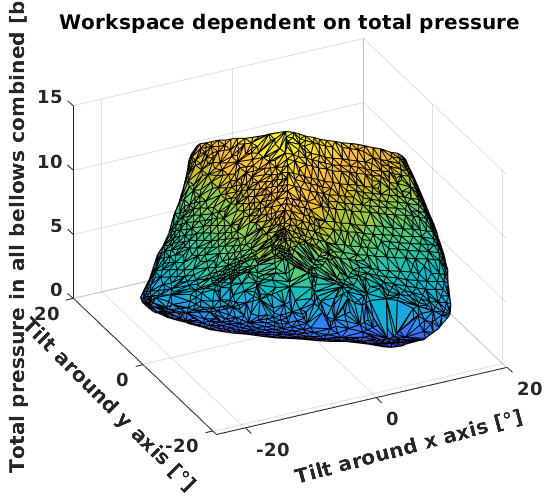}
    \caption{Aggregate pressure augmented workspace}
    \label{fig:augmentedWorkspace}
\end{figure}

The pointcloud matrix structure is organized as seen in eq. \ref{eq9}
\begin{equation}\label{eq9}
    \textbf{P}_{PC} = [\alpha_x, \alpha_y, P_{1C}, P_{2C}, P_{3C}]
\end{equation}
where $\alpha_x$ and $\alpha_y$ are  the measured stable tilt angles which are the result of corresponding input pressures for the respective bellows $P_{1C}$, $P_{2C}$ and $P_{3C}$.
Because of over actuation and the parallel nature of the module design, one orientation of the module is achievable by an infinite combination of input pressures. This can be seen on Fig. \ref{fig:augmentedWorkspace}. Here the $x$-axis and $y$-axis are the tilt around the respective axis in degrees and the $z$-axis is the aggregate pressure, which is the sum of all bellows input pressures in bars. A higher aggregate pressure corresponds to a higher mechanical stiffness of the system. The control algorithm needs to take this into account.

The experimentally measured data represent points in aggregate pressure augmented workspace. To find the inlet pressures from the measured point-cloud, Algorithm 1 was applied.

\begin{algorithm}[h!]
    \caption{Extract required bellows input pressures from point cloud}\label{alg1}
    
    \textbf{Input:} Reference tilt angles $\alpha_{xref}$, $\alpha_{yref}$, required aggregate pressure level $P_{agr}$ and ${P}_{PC} = [\alpha_x, \alpha_y, P_{1C}, P_{2C}, P_{3C}]$ 
    
    \textbf{Output:} Required bellows input pressures $P_1$, $P_2$ and $P_3$ to reach $\alpha_{xref}$ and $\alpha_{yref}$
    \begin{algorithmic}
        \While{isEmpty(Region)}
            
            Add all points to Region that pass the criterion:

            $\sqrt{(\alpha_{xcloud} - \alpha_{xref})^2} \leq$ anT \& $|P_{1cloud} + P_{2cloud} + P_{3cloud} - P_{agr}| \leq$ aggrT
            \If{isEmpty(Region)}
            
                \State $anT = anT +$ incrementAngle

                \State $aggrT = aggrT +$ incrementPressure
            \EndIf
        \EndWhile

        Find point $\textbf{Q} \in Region$  with minimal expression \\
        $\sqrt{(\alpha_{xregion} - \alpha_{xref})^2 - (\alpha_{yregion} - \alpha_{yref})^2}$

        \If{$\sqrt{(\alpha_{xQ} - \alpha_{xref})^2 - (\alpha_{yQ} - \alpha_{yref})^2} \leq $ acceptableTol}
        
            $[P_1, P_2, P_3] = [P_{1Q}, P_{2Q}, P_{3Q}]$

        \Else
        
            $[P_1, P_2, P_3] = [mean(P_{1region}),  mean(P_{2region}), mean(P_{regionQ})]$
        \EndIf
    \end{algorithmic}
\end{algorithm}

$Region$- matrix of measured input pressures and corresponding tilts that will be used to calculate the ended input pressures to reach desired tilt; $anT$- maximum Euclidean distance of a measured point from the reference point in the augmented workspace in the $\alpha_{x}$ $\alpha_{y}$ plane to be eligible for inclusion in $Region$; $aggrT$- maximum Euclidean distance of a measured point from the reference point in the augmented workspace along the aggregate pressure axis to be eligible for inclusion in $Region$; $incrementAngle$- increment to expand $anT$ in case the previous search yealdet empty $Region$; $incrementPressure$- increment to expand $aggrT$ in case the previous search yealdet empty $Region$

\begin{figure}[h!]
    \centering
    \includegraphics[trim={0cm 0cm 0cm 0cm},clip, width=55mm]{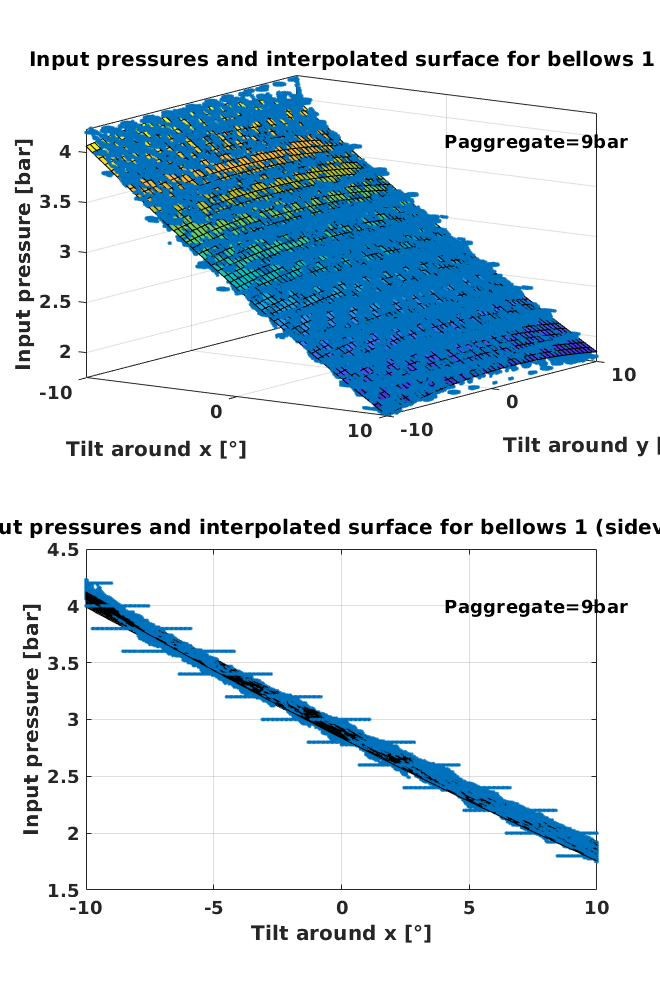}
    \caption{Calculated input pressures and interpolated surface for bellow 1}
    \label{fig:interpolatedSurface}
\end{figure}

Alg. \ref{alg1} will already supply a set of usable input pressures. Unfortunately, the results are influenced by errors in measurement and effects of hysteresis. In a smooth trajectory tracking task this can produce erratic, non-smooth input pressures. To solve these issues, the above Alg. \ref{alg1} was supplied with a set of reference angles ranging from -10° to 10° with an increment of 0.05° for both $\alpha_x$ and $\alpha_y$ and a constant aggregate pressure $P_{agr}$. The result are three 3D meshes representing the relationship between the reference angles and the three input pressures separately. These meshes were then separately interpolated as a surface using a second order $x$ and second order $y$ surface plot. The result for input pressure 1 can be seen in Fig. \ref{fig:interpolatedSurface} and the equation describing this surface is eq. \ref{eq10}

\begin{equation}\label{eq10}
    \begin{split}
    & P_1 = 2.964 - 0.1113\alpha_x + 0.000344\alpha_y + 0.000726\alpha_{x}^2 + \\
    & 0.00407\alpha_x\alpha_y - 0.00123\alpha_{y}^2
    \end{split}
\end{equation}

\begin{figure}[h!]
    \centering
    \includegraphics[trim={0cm 0cm 0cm 0cm},clip, width=85mm]{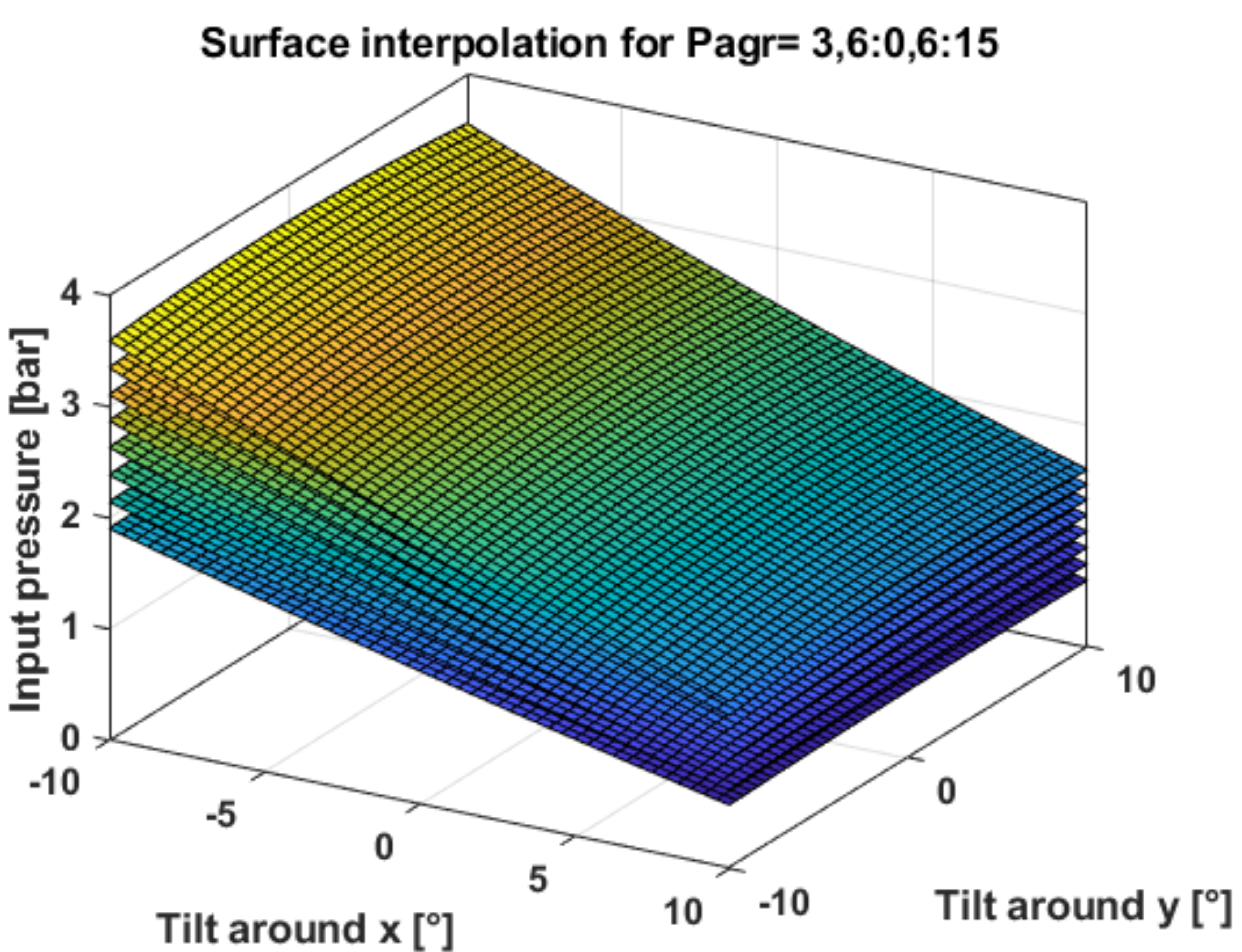}
    \caption{Surface interpolation for $P_{agr}$ = 3.6:0.6:15 for bellows 1}
    \label{fig:surfaceInterpolation}
\end{figure}

\begin{figure}[h!]
    \centering
    \includegraphics[trim={0cm 0cm 0cm 0cm},clip, width=85mm]{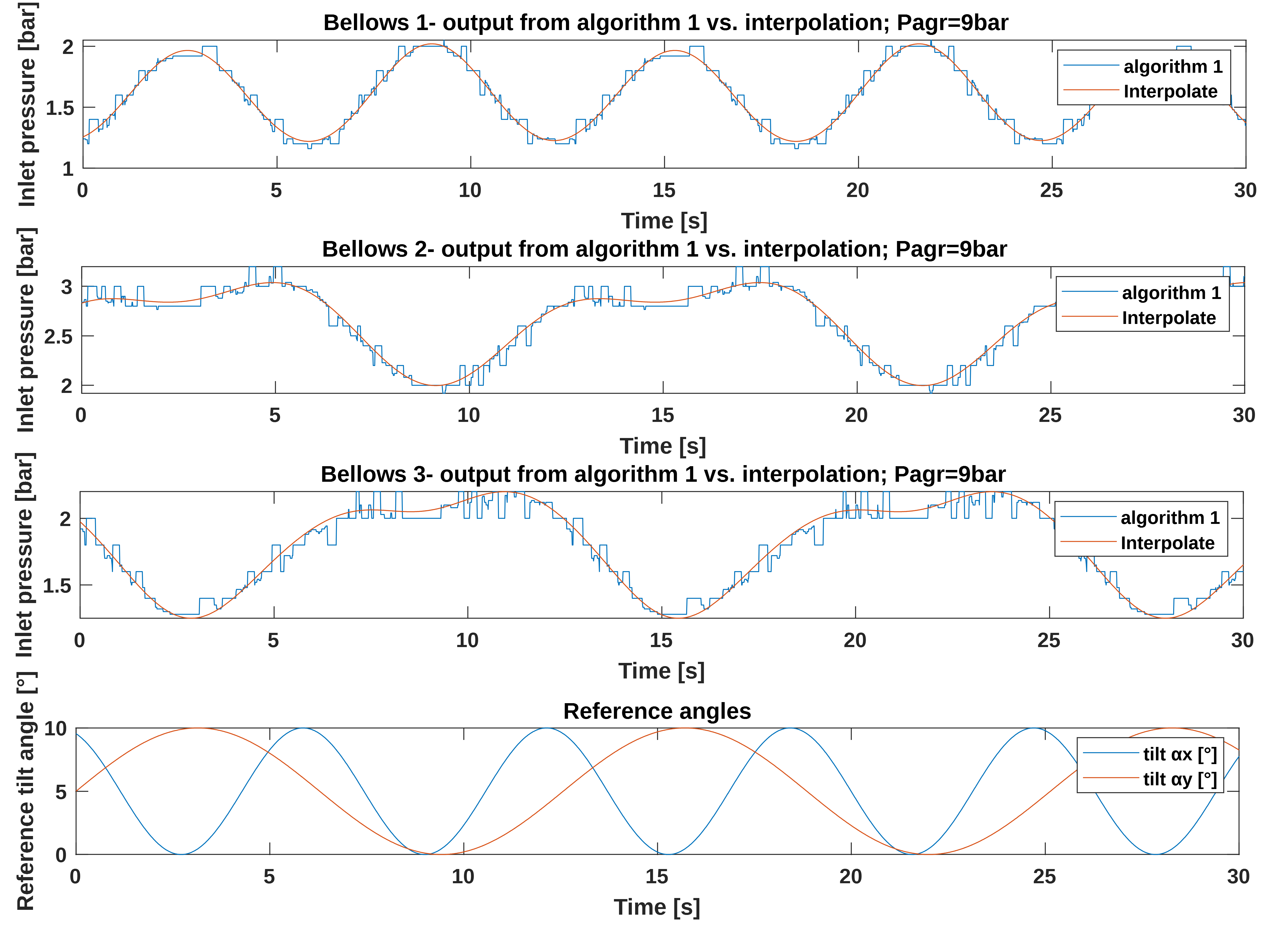}
    \caption{Output from algorithm 1 vs. Interpolation}
    \label{fig:algorithmVSinterpolation}
\end{figure}

This process was repeated for aggregate pressures between 3.6 bar and 15 bar with increment of 0.6 bar for all three bellows. The result is a set of smooth surfaces representing the complete augmented workspace, see Fig. \ref{fig:surfaceInterpolation} for bellows 1. 

The coefficients describing all the surfaces can be further interpolated to get six equations approximating the complete aggregate pressure augmented workspace. The coefficients were interpolated by a 7-th order polynomial. Fig. \ref{fig:algorithmVSinterpolation} compares the output of Alg. \ref{alg1} and the interpolated feed-forward controller. The output of Alg. \ref{alg1} follows the smooth reference signal, but it shows non-smooth, erratic step behavior, while the output of the interpolated feed-forward controller is smooth.

\subsection{Variable gain I-controller design}
\begin{figure}[H]
    \centering
    \includegraphics[trim={0cm 0cm 0cm 0cm},clip, width=85mm]{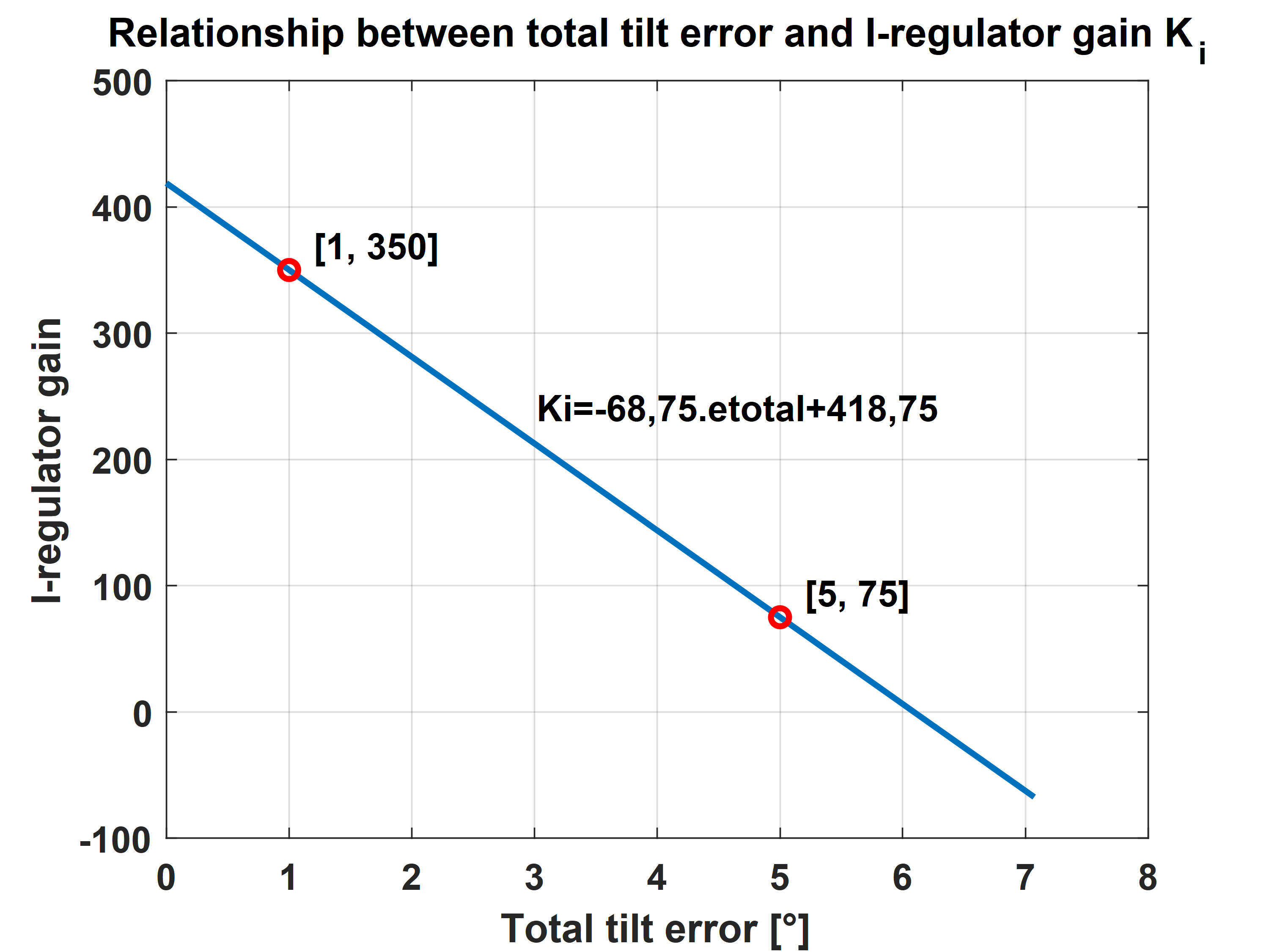}
    \caption{Tilt error in dependence on I-controller gain}
    \label{fig:tiltError}
\end{figure}

The goal of adding a I-part to the controller design is to facilitate disturbance rejection by integrating the reference error over time and scaling it by using a gain. In a constant gain I-controller the gain is tuned to and fixed at a value dependent on the controlled plant. It is simple, easy to implement and does not necessarily require the plant model for correct design and unlike a proportional controller it allows for complete error compensation in a step response. The problem with using a constant gain I-controller is as follows. The feed-forward controller can immediately supply a rough estimate for input pressures, but it is never clear before the movement ends how good this estimation is. Therefore, if the estimate is optimal, a constant gain I-controller would cause overshoot, requiring the controller to be slow. On the other hand, if the estimation is sub-optimal, an aggressive constant gain I-controller is needed to quickly compensate the error. This contradiction in requirements can be solved by applying a variable gain I-controller, with the gain dependent on the error, see Eq. \ref{eq11}
\begin{equation}\label{eq11}
    u(t) = K_i\Bigl(e(t)\Bigr) \int_{0}^{t} e(t) \,dt 
\end{equation}
where $K_i\Bigl(e(t)\Bigr)$ is controller gain, $e(t)$ is tilt error, $t$ is time and $u(t)$ is controller output. The relationship between I-controller gain and tilt error can be described by different types of smooth monotonic functions, like linear, exponential etc. For this controller, as a proof of concept a linear relationship was chosen, see Eq. \ref{eq12}.  
\begin{equation}\label{eq12}
    K_i = ae_t(t) + b
\end{equation}

\begin{equation}\label{eq13}
    e_t(t) = \sqrt{e_{x}^2(t) + e_{y}^2(t)}
\end{equation}
where $e_t(t)$ is total tilt error, $e_x(t)$ is tilt error around $x$-axis and $e_y(t)$ is tilt error around $y$-axis. Parameters $a$ and $b$ were calculated from experimental data, where $K_i = 350$ was found to work well for small total error values below 1° and $K_i = 75$ was found to not cause significant overshoot at error values bellow 5°. This relationship is described by eq. \ref{eq14} and can be seen on Fig. \ref{fig:tiltError}
\begin{equation}\label{eq14}
    K_i = -68.75e_t + 418.75
\end{equation}

\begin{figure}[H]
    \includegraphics[trim={0cm 0cm 0cm 0cm},clip, width=85mm]{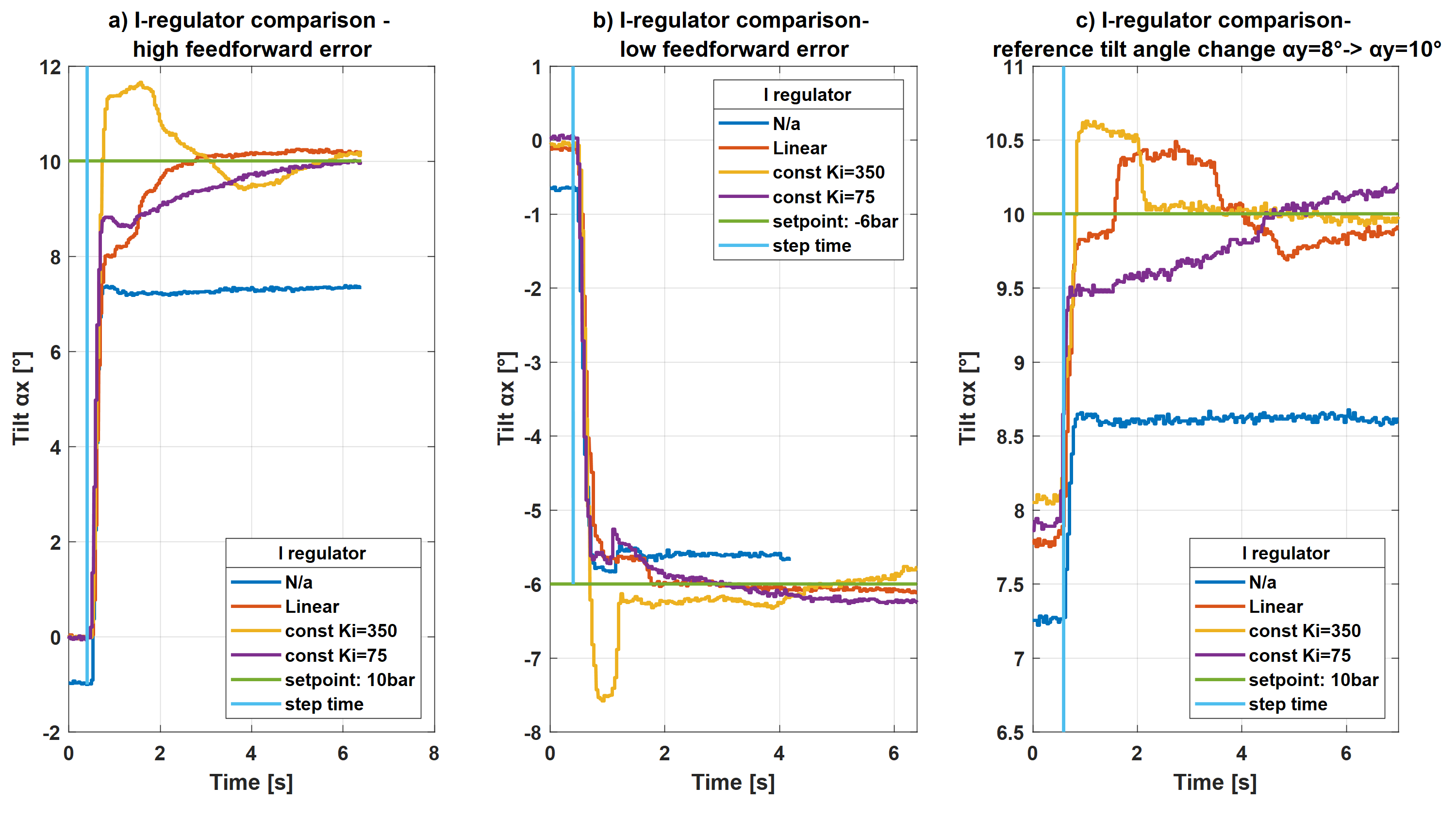}
    \centering
    \caption{Comparison between constant and variable gain I-controller under different conditions}
    \label{fig:IcontrollerComparison}
\end{figure}

The comparison between the performance of a constant gain I-controller and our variable gain controller in combination with our feed-forward controller is depicted in Fig. \ref{fig:IcontrollerComparison}. One can see, that in cases a) and b) a gain of 350 results in a significant overshoot, but for small changes in tilt and a large residual error after feed-forward controller action, like in case c), the controller is fast and has acceptable overshoot. On the other hand, a gain of 75 has no overshoot in any case but is slow and has best performance if the residual error from feed-forward controller action is small, like in case b). The performance of both fixed gain controllers in case c) is not satisfactory. Our variable gain controller performs satisfactory in all cases.

As mentioned above, for our controller, we have chosen a linear relationship to govern the variable gain. A linear relationship is simple, is easy to tune and should result in predictable behaviour of the module. Nevertheless, other types of functions like exponential, polynomial or even logarithmic can also be applied and could result in better controller performance than with a linear equation. The purpose of this article is to show the viability of the by us presented controller and using a linear relationship for this purpose is sufficient. The complexity of comparing different types of governing equations and the required tuning methods warrant its own  article.

\section{Experimental verification of FFvI controller}
This section will focus on experimental comparison between FFvI controller and controllers designed according to established algorithms. First of all, the performance of the feed-forward part of the FFvI algorithm will be compared with a ANFIS controller designed using the same dataset as our feed-forward controller. In the second part the complete FFvI controller will be compared to a constant gain PID controller.

The ANFIS controller was designed using the MATLAB neuro-fuzzy designer. Three controllers for each bellows separately were created. The input data are tilt angles $\alpha_x$ and $\alpha_y$ and the output is the corresponding pressure. The teaching data are picked from the same data, that is used to design the feed-forward controller, but is limited to having an aggregate pressure of $9\pm{1.5}$ bar. The feed-forward controller is also set to the same level of aggregate pressure. This will decrease the teaching time and ensure more reliable results. The neural network used is a Sugeno type network \cite{Takagi1985FuzzyIO} with 10 linear generalized bell-shaped membership functions for each input. The minimum achieved teaching error is in Tab. \ref{tab:tab2}.

\begin{table}[H]
    \caption{Minimum teaching error}
    \centering
    \begin{tabular}{|| c c ||}
    \hline
        Bellows & Teaching error\\
        \hline\hline
        1 & 0.0205\\
        2 & 0.0227\\
        3 & 0.0272\\
        \hline
    \end{tabular}
    \label{tab:tab2}
\end{table}

\begin{figure}[H]
    \centering
    \includegraphics[trim={0cm 0cm 0cm 0cm},clip, width=55mm]{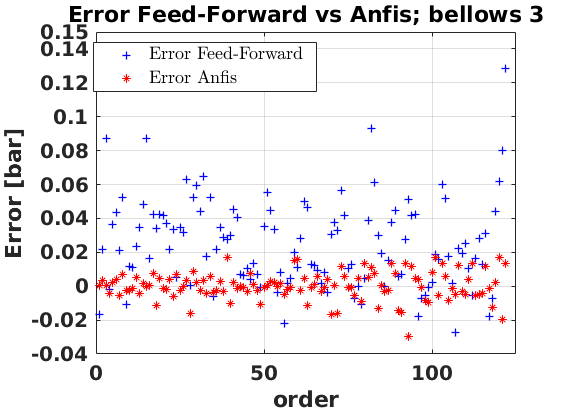}
    \caption{Comparison between feed-forward and ANFIS controller}
    \label{fig:anfis3}
\end{figure}

Both controllers were supplied with the sets of reference tilts and their outputs were compared to input pressures creating the reference tilts. The resulting comparison between the outputs of the feed-forward controller and the ANFIS controller are shown in Fig. \ref{fig:anfis3}. The minimum error and mean error are in Tab. \ref{tab:tab3}.

\begin{figure}[H]
    \centering
    \includegraphics[trim={0cm 0cm 0cm 0cm},clip, width=55mm]{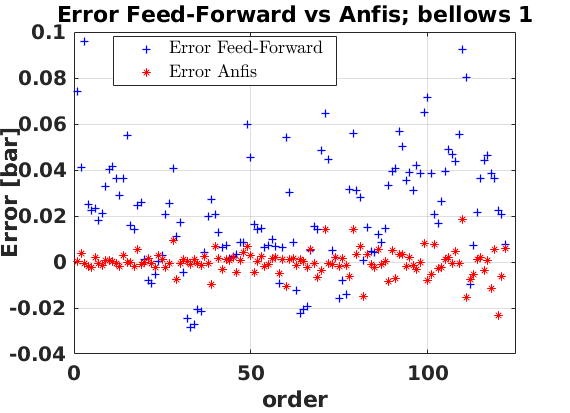}
    \label{fig:anfis1}
\end{figure}

\begin{figure}[H]
    \centering
    \includegraphics[trim={0cm 0cm 0cm 0cm},clip, width=55mm]{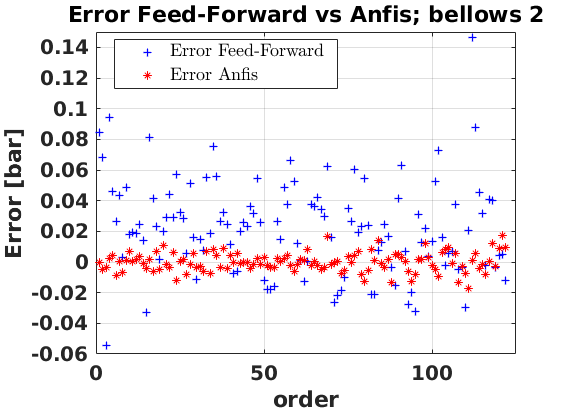}
    \label{fig:anfis2}
\end{figure}

\begin{table}[H]
    \caption{Maximum error}
    \centering
    \begin{tabular}{|| c c c c ||}
    \hline
     Bellows & Controller & maximum error & mean error \\ 
     \hline\hline
     \multirowcell{2}{1} & {feed-forward} & {0.0959} & {0.0217}\\
                      &  {ANFIS} & {0.0231} & {-0.0028} \\
                      \hline
     \multirowcell{2}{2} & {feed-forward} & {0.1463} & {0.0219}\\
                      &  {ANFIS} & {0.0174} & {-0.0011} \\      
                      \hline
    \multirowcell{2}{2} & {feed-forward} & {0.1286} & {0.025}\\
                      &  {ANFIS} & {0.0294} & {0.0029} \\ 
    \hline
    \end{tabular}
    \label{tab:tab3}
\end{table}

The ANFIS controller has a consistently lower error and the mean error is also lower. Newer the less, as can be seen from the above results, both controllers perform well, with the maximum error of the feed-forward controller not exceeding 0.15°. The magnitude of this error is still well within what would be considerate acceptable for this application. Considering that the feed-forward controller can approximate the whole pressure augmented workspace while the ANFIS controller is specifically designed to work in the tested aggregate pressure region, these results are encouraging. 

When assessing the complete FFvI controller, one must keep in mind that the physical module is meant to be part of a robot. Therefore, we have specified that the overshoot, when performing a movement, should not exceed 2° on $\alpha_x$ and $\alpha_y$ separately. This number, although arbitrary, should be a good controller benchmark for this case and the controller is expected to be further tuned after being installed in the complete robot control system. 

\begin{figure}[H]
    \centering
    \includegraphics[trim={0cm 0cm 0cm 0cm},clip, width=55mm]{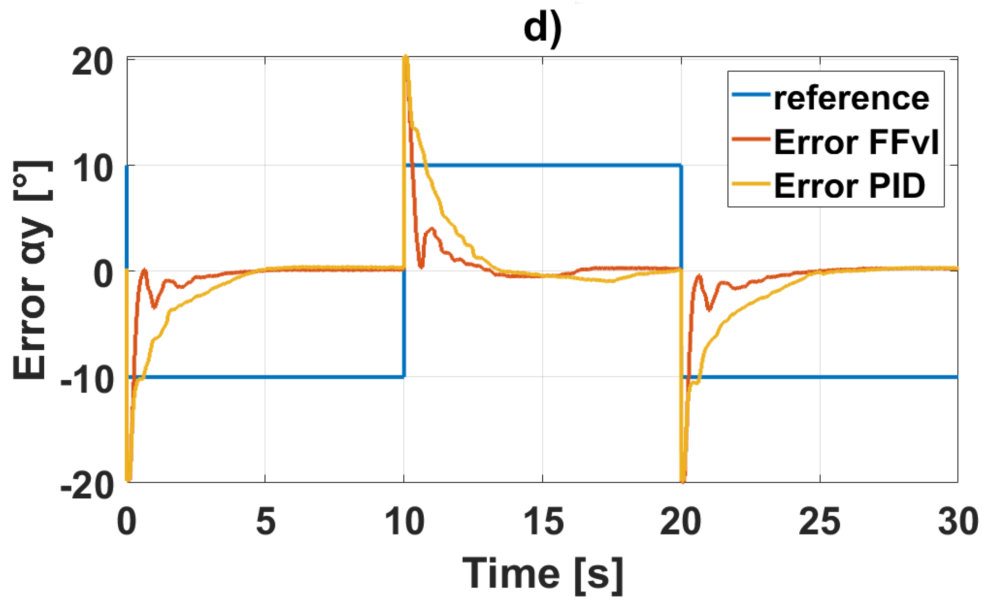}
    \caption{Comparison PID vs. FFvI controller - input step $\alpha_{xref} = -8^\circ$ to $8^\circ$; $\alpha_{yref} = -10^\circ$ to $10^\circ$}
    \label{fig:pidVSFFvI_d}
\end{figure}

As a comparison to our FFvI controller a PID-based controller was used. The controller consists of three identical constant gain PID controller controlling all the bellows separately. The PID controllers were tuned using the P-I-D tuning approach described in \cite{haugen2004pid}. The relevant constants can be seen in Tab. \ref{tab:tab4}. Although the values of the tuned constants are not optimal, the authors believe that they are close enough to the optimal values combining adequate speed while not exceeding our criterion of 2° overshoot on  $\alpha_x$ and $\alpha_y$.
\begin{table}[H]
    \caption{PID controller gains}
    \centering
    \begin{tabular}{|| c c ||}
     \hline
     & Gain \\ 
     \hline\hline
     $K_{P}$ & 40\\
     $K_{I}$ & 240\\
     $K_{D}$ & 5\\
    \hline
    \end{tabular}
    \label{tab:tab4}
\end{table}
The performance of both controllers can be seen on Fig. \ref{fig:pidVSFFvI_d}. The reference signal is sequentially alternating between $\alpha_{xref} =8^\circ$, $\alpha_{yref} =-10^\circ$, $\alpha_{xref} = -8^\circ$ and $\alpha_{yref} = 10^\circ$. These values were chosen because all bellows need to be engaged simultaneously to different degrees, they represent values in the middle part of the module workspace and it is expected that most movement will be in this area and they also demonstrate the asymmetric behavior of the platform.

\begin{figure}[H]
    \centering
    \includegraphics[trim={0cm 0cm 0cm 0cm},clip, width=55mm]{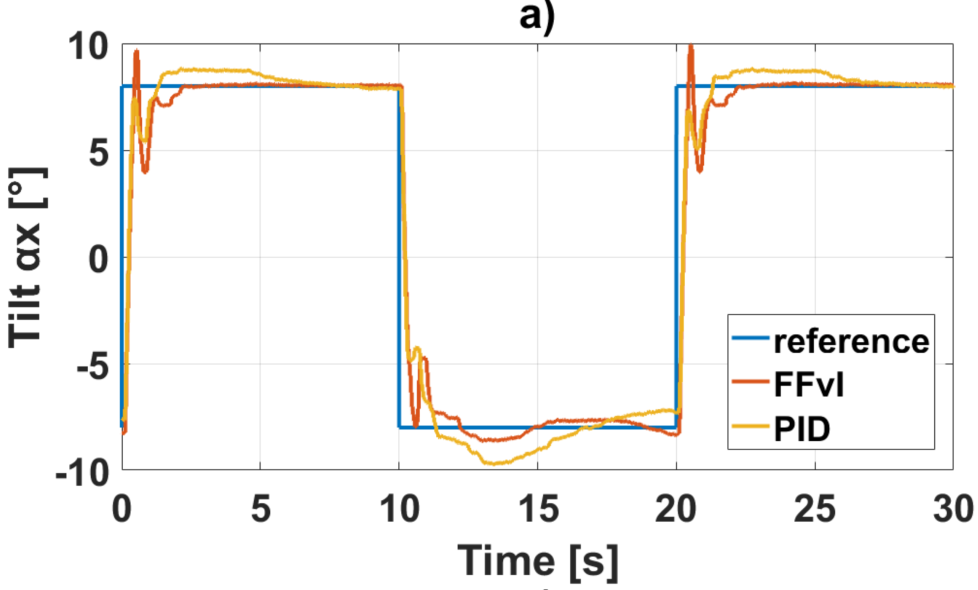}
    \label{fig:pidVSFFvI_a}
\end{figure}
\begin{figure}[h!]
    \centering
    \includegraphics[trim={0cm 0cm 0cm 0cm},clip, width=55mm]{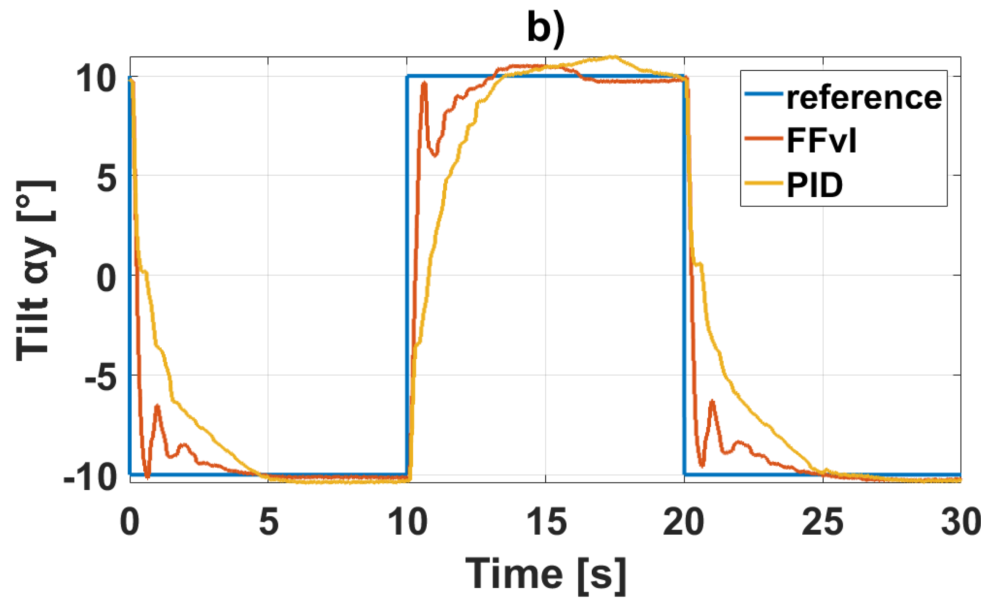}
    \label{fig:pidVSFFvI_b}
\end{figure}
\begin{figure}[H]
    \centering
    \includegraphics[trim={0cm 0cm 0cm 0cm},clip, width=55mm]{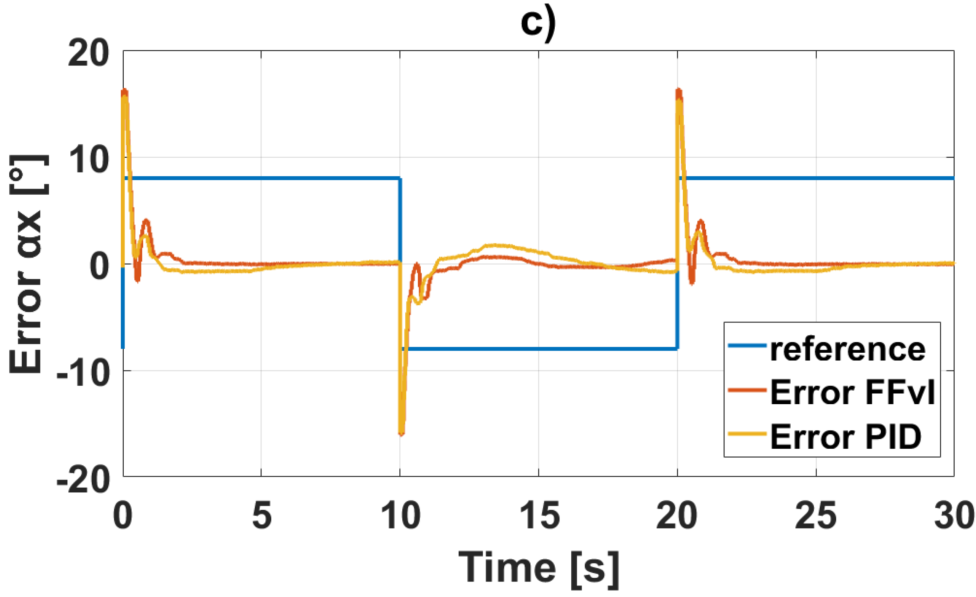}
    \label{fig:pidVSFFvI_c}
\end{figure}

The performance of both controllers types is shown in Tab. \ref{tab:tab5}. It needs to be noted that, this comparison deviates from a standard comparison of step responses by not having the platform at zero tilt and testing a combined rotation around $\alpha_x$ and $\alpha_y$. Nevertheless, this comparison is much closer to comparing the controller performance under more realistic conditions.

\begin{table}[h!]
    \caption{Performance comparison}
    \centering
        \begin{tabular}{|| c c c ||}
             \hline
             $\alpha_x -8 \rightarrow 8$ & FFvI & PID \\ 
             \hline\hline
             rise time [s] & 0.42 & 0.4\\
             setling time 5\% [s] & 2.03 & 5.42\\
             first maximum [s] & 0.54 & 2.15\\
             overshoot [\%] & 21.1 & 10.3\\
            \hline
        \end{tabular}
        
        \begin{tabular}{|| c c c ||}
             \hline
             $\alpha_y 10 \rightarrow -10$ & FFvI & PID \\ 
             \hline\hline
             rise time [s] & 2.74 & 4.13\\
             setling time 5\% [s] & 3.43 & 4.5\\
             first maximum [s] & 0.65 & 7.45\\
             overshoot [\%] & 3 & 3\\
            \hline
        \end{tabular}
\end{table}

\begin{table}[h!]
    
    \centering
        \begin{tabular}{|| c c c ||}
             \hline
             $\alpha_x 8 \rightarrow -8$ & FFvI & PID \\ 
             \hline\hline
             rise time [s] & 0.5 & 1.23\\
             setling time 5\% [s] & 4.46 & 5<\\
             first maximum [s] & 0.57 & 3.4\\
             overshoot [\%] & 8 & 21.3\\
            \hline
        \end{tabular}

        \begin{tabular}{|| c c c ||}
             \hline
             $\alpha_y -10 \rightarrow 10$ & FFvI & PID \\ 
             \hline\hline
             rise time [s] & 0.53 & 3\\
             setling time 5\% [s] & 2.73 & 8.04\\
             first maximum [s] & 0.64 & 7.5\\
             overshoot [\%] & 5 & 9.6\\
            \hline
        \end{tabular}
    
    \label{tab:tab5}
\end{table}

\begin{figure}[h!]
    \centering
    \includegraphics[trim={0cm 0cm 0cm 0cm},clip, width=55mm]{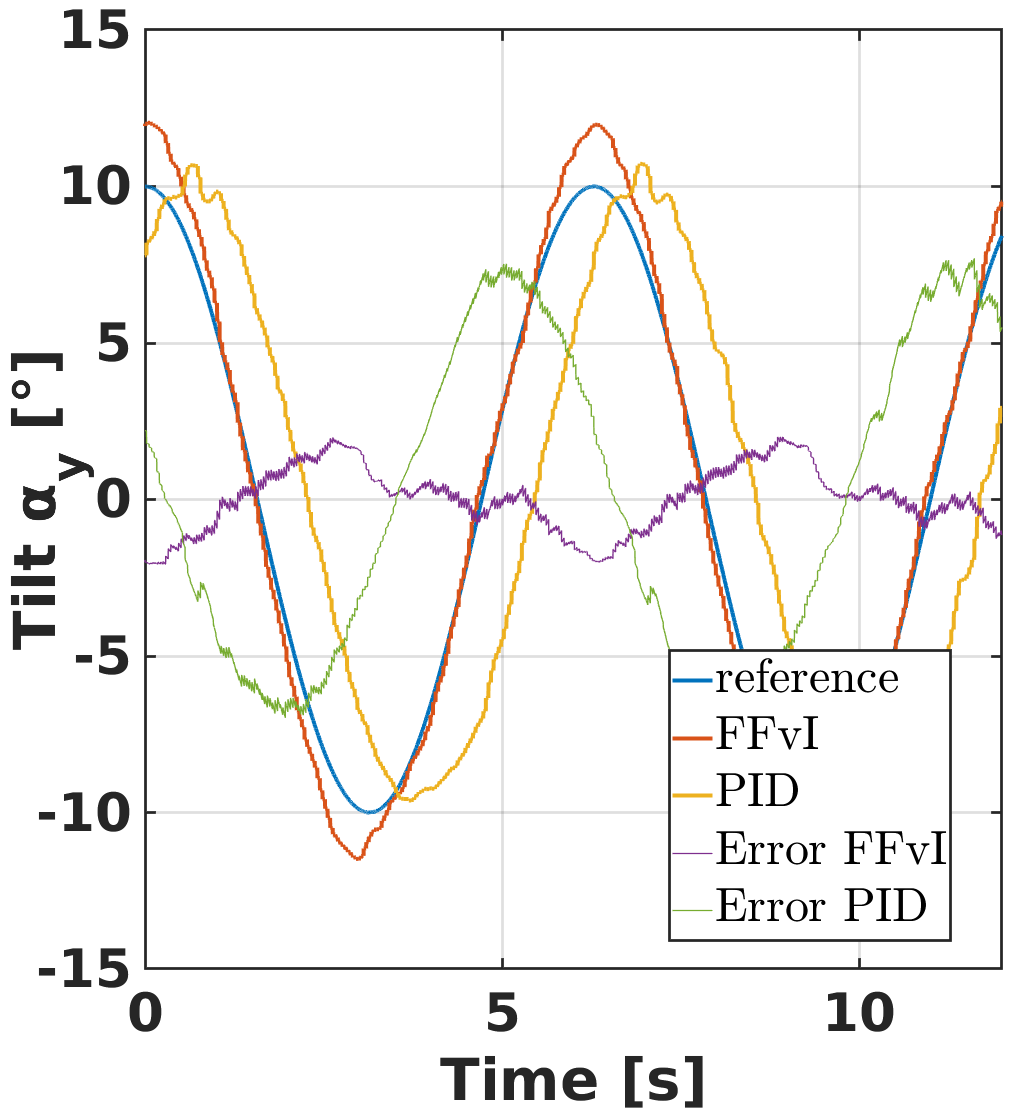}
    \caption{Comparison PID controller vs. FFvI controller - input sinus wave}
    \label{fig:sineY}
\end{figure}

\begin{table}[h!]
    \caption{Sinus reference signal parameters}
    \centering
    \begin{tabular}{|| c c c ||}
    \hline
         sinus signal & $\alpha_{xref}$ & $\alpha_{yref}$\\
         \hline\hline
         amplitude [°] & 10 & 10\\
         period [s] & $2\pi$ & $2\pi$\\
         phase [rad] & 0 & $\pi/2$\\
         bias [°] & 0 & 0\\
    \hline
    \end{tabular}
    \label{tab:tab6}
\end{table}
It can be seen, that both controllers fulfill the condition of not having more than $2^\circ$ overshoot. From Tab. \ref{tab:tab5} it can be seen, that the FFvI controller is faster in all categories and has more overshoot only on when going from $-8^\circ \rightarrow 8^\circ$. 
The second comparison between both controllers is in following a sine reference signal with parameters written in Tab. \ref{tab:tab6}, as can be seen in Fig. \ref{fig:sineY}. 
Here the FFvI controller can tightly and smoothly follow the sinus reference signal while overshooting at the maximum and minimum of the reference signal. The overshoot, again, does not exceed 2° for either tilt angles. The PID controller, while not overshooting nearly as much, lags constantly behind the reference signal by about 0.7 s and the plot is in some parts jittery. The overall tracking error is much smaller for FFvI than for PID controller. It can be said that both controllers perform acceptably, while the FFvI controller is faster while still passing the overshoot criteria. 

\begin{figure}[h!]
    \centering
    \includegraphics[trim={0cm 0cm 0cm 0cm},clip, width=55mm]{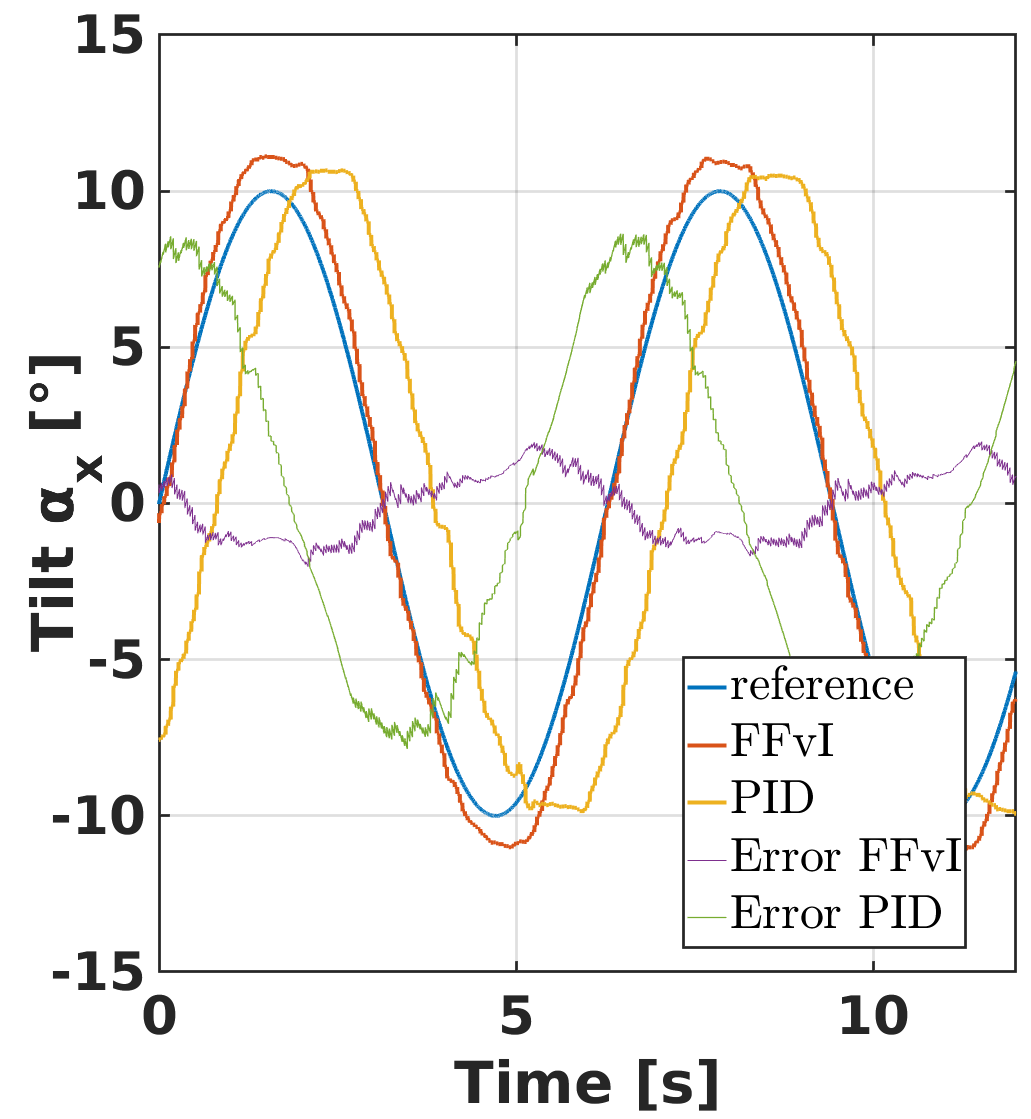}
    \label{fig:sineX}
\end{figure}

\begin{figure}[h!]
    \centering
    \includegraphics[trim={0cm 0cm 0cm 0cm},clip, width=55mm]{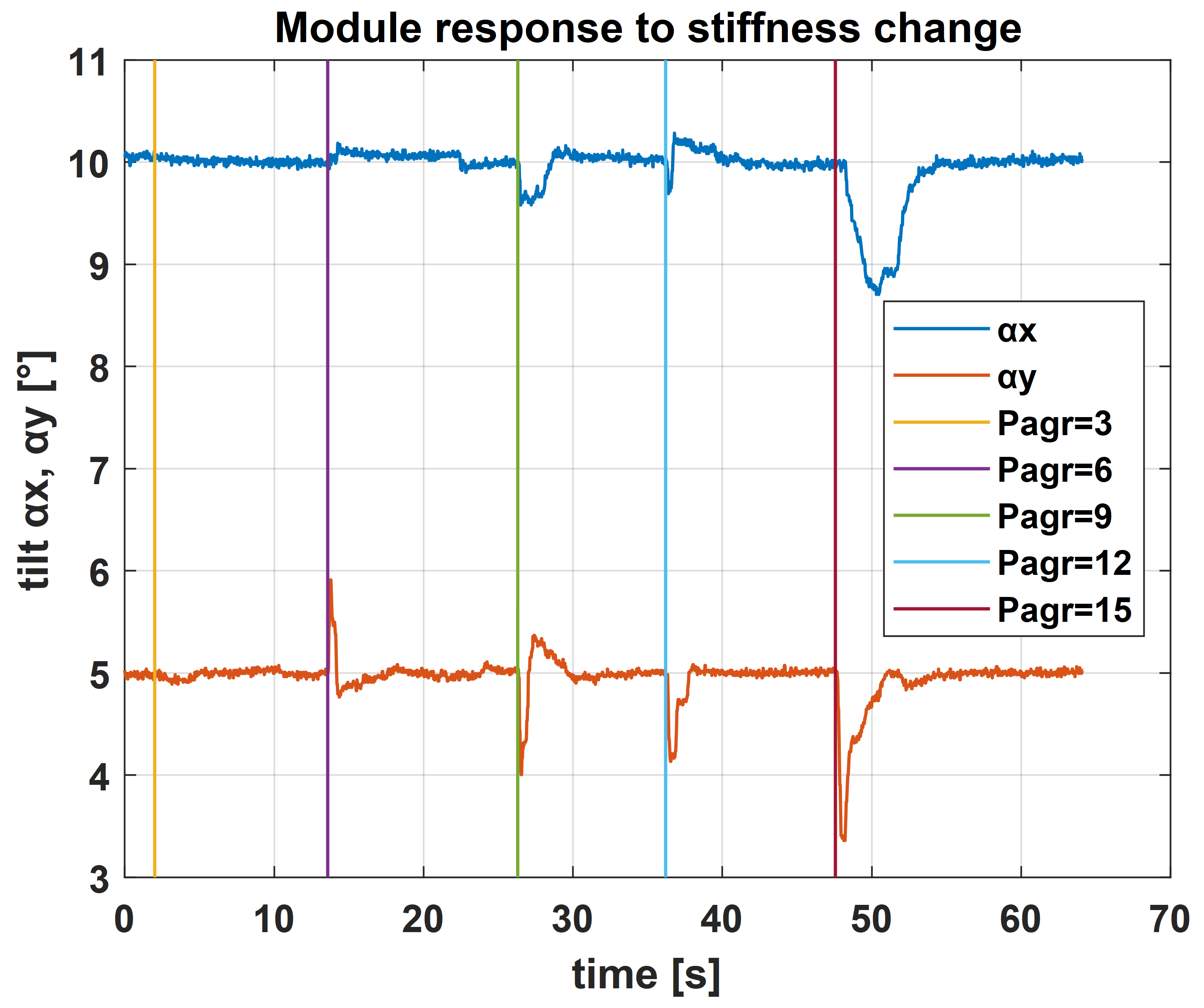}
    \caption{Module response to stiffness change}
    \label{fig:stiffness}
\end{figure}

One ability that the FFvI controller has is to change the stiffness of the system while maintaining reference tilt. This is shown in Fig. \ref{fig:stiffness}. The module is first tilted to $\alpha_{xref} = 10^\circ$, $\alpha_{yref} = 5^\circ$ at a requested aggregate pressure of 3 bar. Then, the aggregate pressure is changed stepwise 6 bar, 9 bar, 12 bar and 15 bar, respectively. A change in aggregate pressure corresponds to a change in stiffness of the system. As can be seen from the Fig. \ref{fig:stiffness} pressure and therefore stiffness can be changed online. This change results in a momentous destabilization of the system resulting in slight position loss. The maximum error for our test was at the transition between aggregate pressure 12 bar and 15 bar with maximum error for $\alpha_{xref} = 1.3^\circ$, $\alpha_{yref} = 1.65^\circ$. This can be again, attributed to nonsynchronous pressure change between the bellows and possible measurement errors in the original feed-forward controller input data. Therefore, instead of a sharp aggregate pressure step a smooth aggregate pressure transition should decrease this issue.

\section{Dynamic test of FFvI tilt platform controller}
To be able to apply the proposed controller as part of the control system of the whole manipulator it is necessary to test the controller under dynamic conditions.

\begin{figure}[H]
    \centering
    \includegraphics[width=0.7\linewidth]{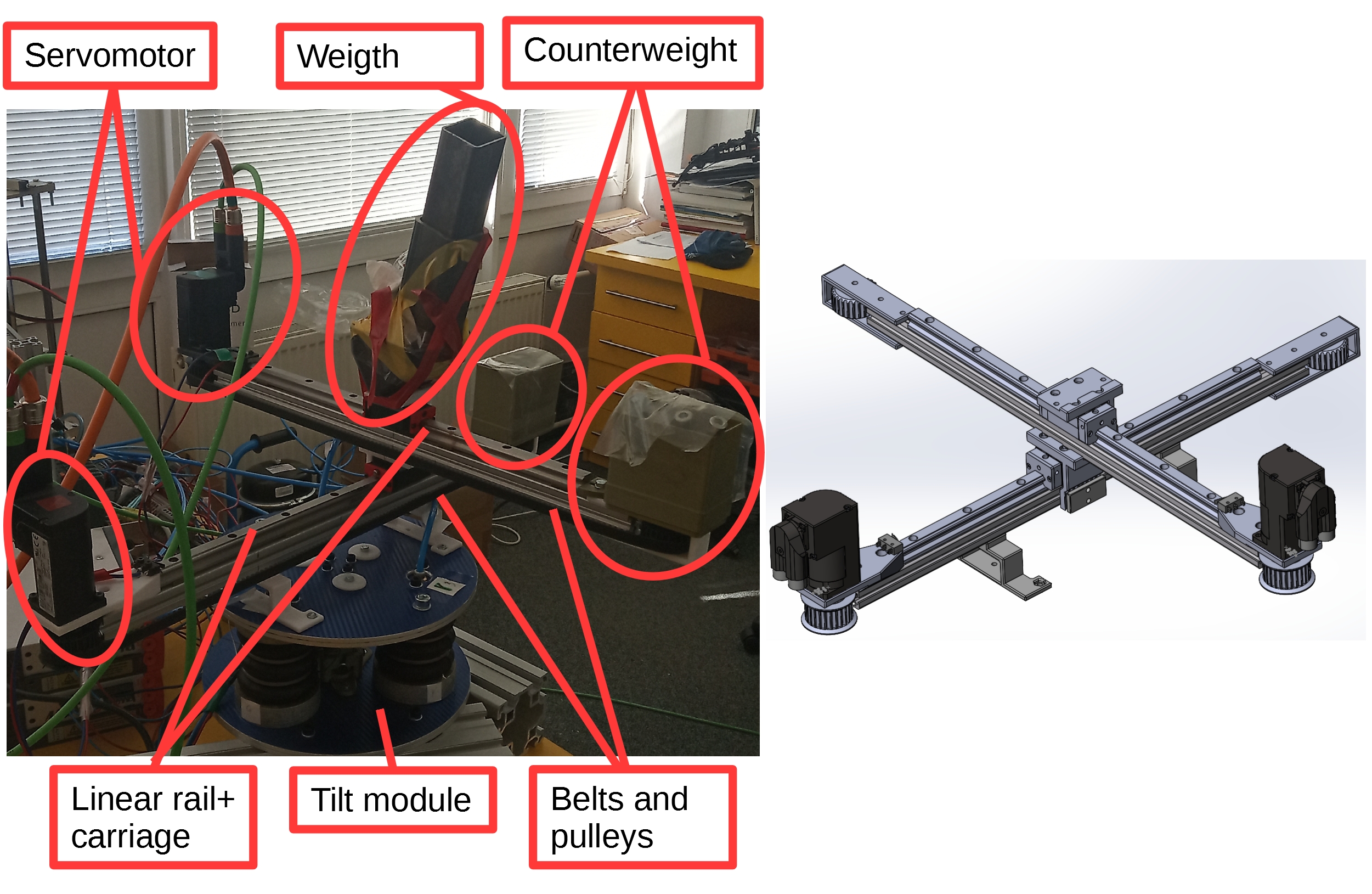}
    \caption{Dynamic two axis loading mechanism}
    \label{fig:loading_device}
\end{figure}

For this purpose a two axis loading mechanism was developed, see Fig.~\ref{fig:loading_device}. It consists of two linear motion axis stacked perpendicular on top of each other and a  attached to the top axis. This mechanism is mounted on top of a tilt module with the axis axis of the tilt platform aligned with the linear motion axis. It can be seen that the potential loading momentum will be different for both axis because the bottom axis is loaded not only by the loading weight but also the top axis itself. The possible motion of one linear axis is 0$mm$ to 330$mm$ and is centered on the central axis of the tilt module. The weight of the axis and loading weight is in Tab.~\ref{tab:weigths} .This mechanism is used to generate dynamic loading forces to study the capability of the FFvI controller to reject dynamic disturbances.

\begin{table}[h!]
    \caption{Loading weigths}
    \centering
    \begin{tabular}{|| c c ||}
     \hline
     & weight [g] \\ 
     \hline\hline
     $linear~axis$ & 2467\\
     $loading~weigth$ & 1802\\
    \hline
    \end{tabular}
    \label{tab:weigths}
\end{table}

\begin{figure}[h!]
     \centering
     \begin{subfigure}[b]{0.38\textwidth}
         \centering
         \includegraphics[width=0.7\linewidth]{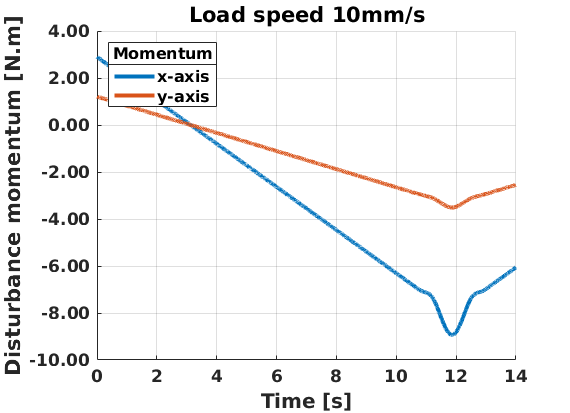}
     \end{subfigure}
     \hspace{3em}%
     \begin{subfigure}[b]{0.38\textwidth}
         \centering
        \includegraphics[width=0.7\linewidth]{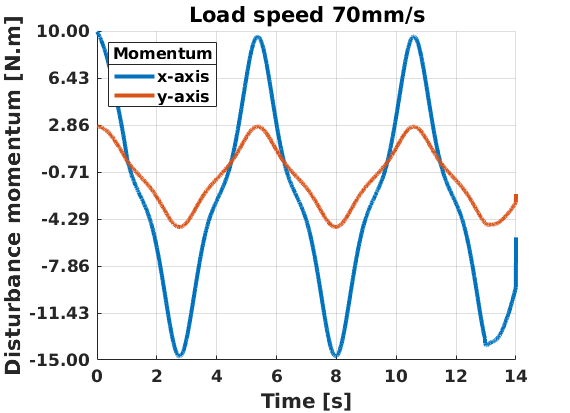}
     \end{subfigure}

    \caption{Disturbance momentum at different loading speeds}
\label{fig:disturbance_momentum}
\end{figure}

\begin{figure}[h!]
     \centering
     \begin{subfigure}[b]{0.390\textwidth}
         \centering
         \includegraphics[width=0.7\linewidth]{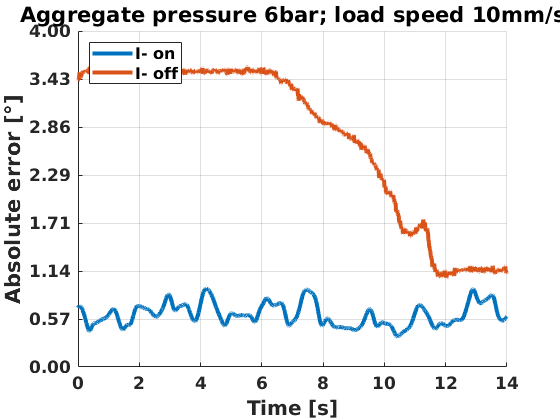}
     \end{subfigure}
     \hspace{3em}%
     \begin{subfigure}[b]{0.39\textwidth}
         \centering
        \includegraphics[width=0.7\linewidth]{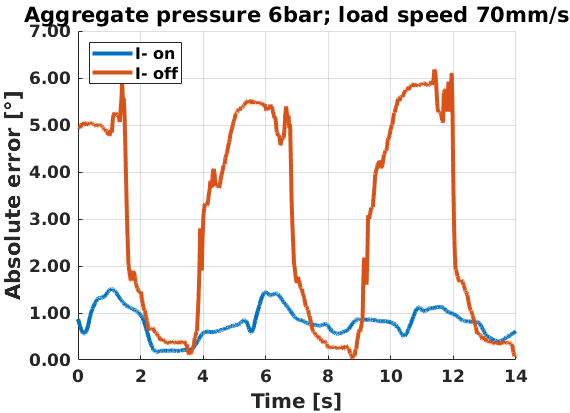}
     \end{subfigure}
     \centering
     \begin{subfigure}[b]{0.390\textwidth}
         \centering
         \includegraphics[width=0.7\linewidth]{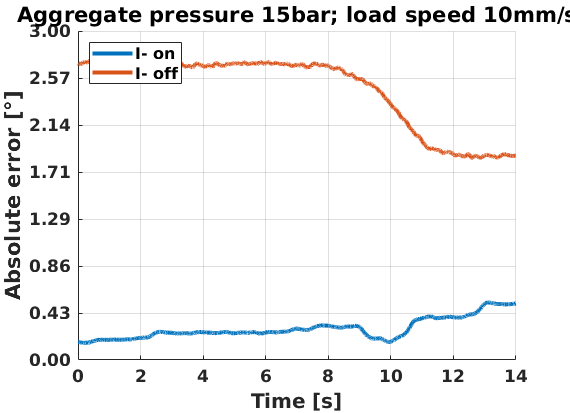}
     \end{subfigure}
     \hspace{3em}%
     \begin{subfigure}[b]{0.39\textwidth}
         \centering
        \includegraphics[width=0.7\linewidth]{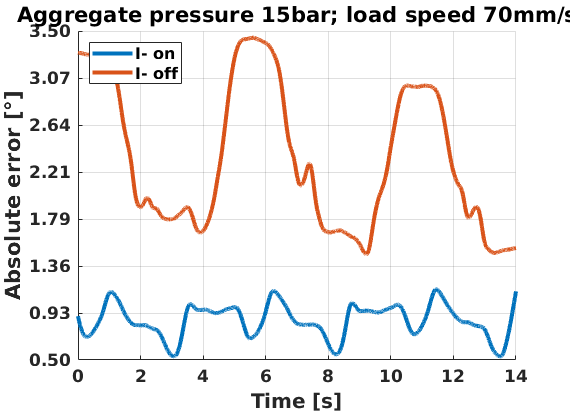}
     \end{subfigure}  
    \caption{Comparison between active and inactive I regulator at tilt holding at extreme aggregate pressures and dynamic effects}
\label{fig:dynam_test}
\end{figure}

The experiment seen in Fig.~\ref{fig:dynam_test} is done for reference tilts $\alpha_x$=0° and $\alpha_y$=0°. Error rejection is facilitated by the I controller part of the algorithm, hence the test will be performed with the I controller active and, for comparison, with the I controller inactive. Also, the experiment will be done for different values of aggregate pressure and for different speeds of movement of the loading weight. The movement of the load was converted to loading momentum around axis $x$ and $y$ (see Fig.~\ref{fig:disturbance_momentum}). Fig.~\ref{fig:disturbance_momentum} and Fig.~\ref{fig:dynam_test} show only the the extremes of the test, different combinations of load speed and aggregate pressure were also tested.

The disturbance momentum was indirectly established from the measured movement of the load and the geometry of the mechanism. The shown time window is the same for all movement speeds. The maximum total error for all measurements is in Tab.~\ref{tab:total_error}.

\begin{table}[h!]
    \caption{Maximum total error under dynamic load}
    \centering
    \begin{tabular}{|| c c c ||}
     \hline
     velocity; pressure & I on [°] & I off [°]\\ 
     \hline\hline
10$mm/s$; 6$bar$ & 0.93	& 3.57\\
30$mm/s$; 6$bar$ & 0.66 & 4.29\\
50$mm/s$; 6$bar$ & 1.52	& 3.48\\
70$mm/s$; 6$bar$ & 1.51	& 6.18\\
10$mm/s$; 9$bar$ & 0.31	& 3.84\\
30$mm/s$; 9$bar$ & 0.62	& 3.75\\
50$mm/s$; 9$bar$ & 1.02	& 3.66\\
70$mm/s$; 9$bar$ & 1.05	& 6.82\\
10$mm/s$; 12$bar$ & 0.28	& 2.78\\
30$mm/s$; 12$bar$ & 0.60	& 2.58\\
50$mm/s$; 12$bar$ & 0.88	& 3.17\\
70$mm/s$; 12$bar$ & 1.15	& 5.42\\
10$mm/s$; 15$bar$ & 0.52	& 2.74\\
30$mm/s$; 15$bar$ & 0.63	& 2.68\\
50$mm/s$; 15$bar$ & 0.87	& 3.24\\
70$mm/s$; 15$bar$ & 1.14	& 3.44\\
mean & 0.86 & 3.85 \\
     
    \hline
    \end{tabular}
    \label{tab:total_error}
\end{table}

From the above figures one can deduce important findings regarding the behaviour of the regulator under dynamic changing load. The controller greatly benefits the precision of the control by rejecting the residual error created by feed-forward error imperfection where the maximum total error with the I part active was 1.52° and the mean error was 0.86° and the maximum total error with the I part inactive was 6.85° and the mean error was 3.85°. Apart from that a well tuned variable gain I-controller has a significant stabilisation effect at higher dynamic loads. It can also be seen, especially at higher speeds of the load, that a higher aggregate pressure has a positive effect on the rejection of dynamic loads.

It can be concluded that his controller in its current state is able to control the prototype manipulator, especially at lower speeds and higher aggregate pressure settings.

\section{Conclusion}
In this paper a new type of controller for the control of the tilt of a pneumatic bellows actuated module of the cascade robot PneuTrunk was presented. The controller consists of a feed-forward controller designed using experimental data and a variable gain I-controller. The feed-forward controller is created by fitting the data at a certain pressure level using a polynomial function and subsequently, again fitting the resulting set of polynomial function constants by another set of polynomials. This allows for a simple and fast controller, that not only allows to control the tilt of the module, but also its stiffness on demand. The variable gain I-controller supplements the feed-forward controller by adding a feedback loop, hence facilitating disturbance rejection and correcting for feed-forward controller imperfections. The variable gain allows for fast error correction while limiting overshoot and windup at the same time. This hybridisation approach allows for a simple controller design for a complex MIMO systems that can be easily adjusted and updated.

This controller was compared to other established controllers. On the feed-forward level, the controller was compared with an ANFIS controller, delivering comparable results. Comparing the complete controller to a tuned PID controller showed that our  FFvI controller is faster, can reliably follow a harmonic reference signal while maintaining required performance parameters. This controller was also testet under dynamic load to satisfactory results. The maximum FFvI controller error during the positioning of the module with consideration of dynamic disturbance was only 1.52°.

To further improve the performance of the controller, it is necessary to create a comprehensive mathematical model of the module, mainly to combat the detrimental effects of the hysteretic behavior of the bellows. It is also appropriate to compare different types of functions driving the variable gain of the I-controller. In the future, this controller will be applied as a part of a larger control system controlling the pneumatic cascade robot PneuTrunk.

\section*{Acknowledgments}
This research was funded by Slovak Grant Agency VEGA 1/0436/22 Research on modeling methods and control algorithms of kinematically redundant mechanisms and VEGA 1/0201/21 Mobile mechatronic assistant. This research has been also elaborated under support of the project Research Centre of Advanced Mechatronic Systems, reg. No. CZ.02.1.01$/0.0/0.0/16_019/$0000867 in the frame of the Operational Program Research, Development and Education.


%
\bibliographystyle{IEEEtran}

\bibliography{main}

\vfill

\end{document}